%% file: acl_latex.tex
\pdfoutput=1

\documentclass[11pt]{article}

\usepackage[final]{acl}

\usepackage{times}
\usepackage{latexsym} 

\usepackage[T1]{fontenc}

\usepackage[utf8]{inputenc}

\usepackage{microtype}

\usepackage{inconsolata}

\usepackage{graphicx}

\usepackage{tcolorbox}
\tcbuselibrary{breakable}

\usepackage{booktabs}
\usepackage{multirow}
\usepackage{subfigure}
\usepackage{wrapfig}
\usepackage{caption}
\usepackage{amsmath,amssymb}
\usepackage{ulem}
\usepackage{colortbl}
\usepackage{arydshln}

\usepackage{subcaption}
\tcbuselibrary{skins}
\usepackage{adjustbox}

\definecolor{mydarkblue}{RGB}{0,0,180}
\usepackage{xcolor}
\usepackage{soul}
\setulcolor{blue}

\usepackage{pifont}

\newcommand{\MODELNAME}{\textsc{BRIEF-Pro}}

\title{\MODELNAME{}: Universal Context Compression with Short-to-Long Synthesis for Fast and Accurate Multi-Hop Reasoning}

\author{
  Jia-Chen Gu\thanks{\hspace{0.5mm}Equal contribution.}, Junyi Zhang\footnotemark[1], Di Wu, Yuankai Li, Kai-Wei Chang, Nanyun Peng  \\
  University of California, Los Angeles \\
  {\tt \{gujc,junyizhang2002\}@ucla.edu}, 
  {\tt \{diwu,kwchang,violetpeng\}@cs.ucla.edu}
}

\begin{document}
\maketitle 
\begin{abstract}
As retrieval-augmented generation (RAG) tackles complex tasks, increasingly expanded contexts offer richer information, but at the cost of higher latency and increased cognitive load on the model. 
To mitigate this bottleneck, especially for intricate multi-hop questions, we introduce \MODELNAME{}. 
It is a universal, lightweight compressor that distills relevant evidence for a given query from retrieved documents into a concise summary for seamless integration into in-context RAG. 
Using seed data consisting of relatively short contexts (fewer than 1k words), \MODELNAME{} is trained to perform abstractive compression of extended contexts exceeding 10k words across a wide range of scenarios.
Furthermore, \MODELNAME{} offers flexible user control over summary length by allowing users to specify the desired number of sentences.
Experiments on four open-domain multi-hop question-answering datasets show that \MODELNAME{} generates more concise and relevant summaries, enhancing performance across small, large, and proprietary language models.
With the 70B reader model, 32× compression by \MODELNAME{} improves QA performance by 4.67\% on average over LongLLMLingua’s 9×, while requiring only 23\% of its computational overhead\footnote{Code and data: \href{https://github.com/JasonForJoy/BRIEF}{https://github.com/JasonForJoy/BRIEF}}.
\end{abstract}

\input{text/1_intro}
\input{text/2_related}

\input{text/3_method}
\input{text/4_experiment}

\input{text/5_conclusion}

\section*{Acknowledgement}
This research was supported in part by NSF \#2331966, Taboola, Amazon and Coefficent Giving.
We would like to express gratitude to the UCLANLP group members for their valuable feedback.

\bibliography{custom}

\clearpage
\appendix
\onecolumn
\input{text/_appendix}

\end{document}

%% file: text/1_intro.tex
\section{Introduction}

\begin{figure}[t]
  \centering
  \includegraphics[width=0.48\textwidth]{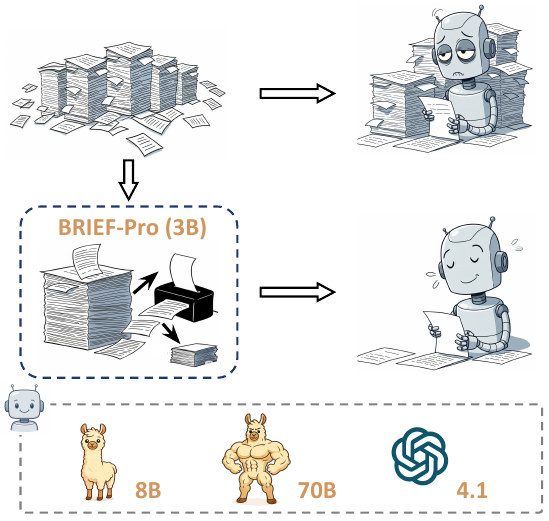}
  \caption{
  A comparison of the inference process with and without the lightweight \MODELNAME{}.
  Retrieved documents are compressed into a highly dense textual summary relevant to the query before being prepended, thereby reducing the cognitive load caused by the extended context on a range of larger models, including 8B, 70B, and proprietary models.  
  }
  \label{fig-inference}
  \vspace{-4mm}
\end{figure}

Retrieval-Augmented Generation (RAG)~\citep{DBLP:conf/nips/LewisPPPKGKLYR020} has emerged as a powerful paradigm for enhancing the factual grounding and knowledge breadth of large language models (LLMs)~\citep{meta2024llama31,anthropic2024claude35,openai2025gpt41,google2025gemini25}. 
By dynamically retrieving relevant information from a vast corpus and incorporating it into the LLM's context, RAG systems aim to mitigate issues like hallucination and outdated knowledge~\citep{DBLP:journals/csur/JiLFYSXIBMF23}.
However, a significant bottleneck in scaling RAG~\citep{DBLP:conf/nips/ShaoHASDMZK24} to real-world applications with numerous retrieved documents is the proportional increase in latency~\citep{DBLP:conf/emnlp/JiangWLYQ23,DBLP:conf/iclr/XuSC24}. 
Feeding such an extensive context not only dramatically slows down inference but also increases the cognitive load imposed on the model~\citep{DBLP:conf/icml/ShiCMSDCSZ23,DBLP:conf/acl/MallenAZDKH23,DBLP:conf/iclr/0008PWM0LSBSC24,DBLP:journals/tacl/LiuLHPBPL24}.
This challenge is amplified for intricate \emph{multi-hop} questions that demand reasoning across multiple disparate sources~\citep{DBLP:conf/emnlp/Yang0ZBCSM18,DBLP:conf/coling/HoNSA20,DBLP:journals/tacl/TrivediBKS22}. 
The large volume of information can overwhelm LLMs, hindering evidence integration and diluting key signals, leading to suboptimal performance even with relevant content.

To address these challenges, context compression has emerged as a crucial technique for improving the efficiency and effectiveness of the RAG system. 
However, existing context compression methods often struggle to retain all critical information when the input context becomes extensive~\citep{DBLP:conf/acl/JiangWL0L0Q24} as illustrated in Appendix~\ref{sec-results}. 
This limitation arises from their inability to capture interdependencies across large or multiple documents.
Moreover, these methods usually do not allow users to set the compression budget~\citep{DBLP:conf/iclr/XuSC24,DBLP:conf/naacl/LiGWCP25}, making it difficult to balance compression rate and information preservation for specific applications.

To mitigate this critical bottleneck and unlock the full potential of RAG for complex reasoning, we introduce improved \MODELNAME{} (Figure~\ref{fig-inference}), a universal, lightweight compressor for \textbf{B}ridging \textbf{R}etrieval and \textbf{I}nference through \textbf{E}vidence \textbf{F}usion. 
\MODELNAME{} is designed to distill relevant evidence for a given query from retrieved documents into a concise summary. 
Unlike previous studies on short-context compression~\citep{DBLP:conf/iclr/XuSC24,DBLP:conf/emnlp/YoonLHJK24,DBLP:conf/naacl/LiGWCP25}, our curated training data enables \MODELNAME{} to generalize to much longer inputs, as required in real-world RAG scenarios.
The key innovation of \MODELNAME{} is its long-context synthetic data pipeline, developed from short-context seed data (fewer than 1k words), enabling abstractive compression of 10k+ word contexts across diverse scenarios. 
Furthermore, \MODELNAME{} allows users to control compression length by designing an instruction-conditioned paradigm, specifying the desired number of sentences to balance conciseness and informativeness for specific application requirements.

We evaluated \MODELNAME{} on four open-domain multi-hop question-answering (QA) datasets: the extended MuSiQue~\citep{DBLP:journals/tacl/TrivediBKS22}, HotpotQA~\citep{DBLP:conf/emnlp/Yang0ZBCSM18}, and 2WikiMultiHopQA~\citep{DBLP:conf/coling/HoNSA20} from LongBench~\citep{DBLP:conf/acl/BaiLZL0HDLZHDTL24}, and LongSeal~\citep{DBLP:journals/corr/abs-2506-01062}, with context lengths ranging from 4.9k to 14.8k words. 
The extensive contexts in these datasets challenge traditional compression methods.
Experimental results consistently demonstrate that \MODELNAME{} generates more concise and relevant summaries. 
This significantly improves the QA accuracy and inference latency across a wide range of small (\texttt{Llama-3.1-8B-Instruct}), large (\texttt{Llama-3.1-70B-Instruct})~\citep{meta2024llama31}, and proprietary (\texttt{GPT-4.1-nano})~\citep{openai2025gpt41} language models. 
Specifically, with the 70B reader model, 32× compression by \MODELNAME{} improves QA performance by 4.67\% on average over LongLLMLingua’s 9×~\citep{DBLP:conf/acl/JiangWL0L0Q24}, while requiring only 23\% of its computational overhead.
These findings highlight \MODELNAME{}'s potential to advance the scalability and effectiveness of RAG for complex retrieval and generation tasks.

In summary, our contributions in this paper are three-fold:
(1) This study pioneers the exploration of multi-hop reasoning and compression of RAG for \textit{long contexts of 10k+ words} across diverse scenarios.
(2) A synthetic data pipeline, built on short-context seed data, is designed to synthesize \textit{long-context} training data for compression learning. 
(3) \MODELNAME{}, trained on the curated dataset, generates concise summaries that \textit{accelerate the inference and enhance the accuracy} of a wide range of small, large, and proprietary language models. 

%% file: text/2_related.tex
\section{Related Work}
The processing and understanding of long contexts presents several challenges, including increased inference costs, longer latency, and decreased performance due to redundant and distracting information.
Many efforts have been made to compress long contexts.
One line of research proposes compressing long contexts into soft prompts that can be used by LMs, such as GIST~\citep{DBLP:conf/nips/Mu0G23}, AutoCompressors~\citep{DBLP:conf/emnlp/ChevalierWAC23}. 
However, these soft prompts are usually tailored to particular tasks and require fine-tuning to align with the representation space of LMs, which severely limits their compatibility.
Another line of work proposes compressing contexts into textual summaries, such as RECOMP~\citep{DBLP:conf/iclr/XuSC24}, CompAct~\cite{DBLP:conf/emnlp/YoonLHJK24}, EXIT~\citep{DBLP:conf/acl/HwangCJSHP25}, and BRIEF~\citep{DBLP:conf/naacl/LiGWCP25}, and our method belongs to this category.
RECOMP distills the summarization ability of extreme-scale proprietary LLMs into an in-house abstractive compressor.
CompAct employs an active strategy to recurrently acquire new information from documents and compress it into a compacted context.
BRIEF curates a synthetic data pipeline built by open-source models to enhance the awareness of multi-hop reasoning and scalability.
Compared to soft prompts, this approach yields more interpretable textual summary that can transfer across different LMs, and can be applied to black-box LMs without requiring gradient updates. 
However, these compression methods are limited to scenarios where the context is relatively short and provide limited user control over compression. 
The LLMLingua family~\citep{DBLP:conf/emnlp/JiangWLYQ23,DBLP:conf/acl/JiangWL0L0Q24} is most relevant to this work, proposing demonstration- and token-level compression that leverages a small LM to calculate perplexity and prune redundancy. 
But their budget allocation may inadvertently discard crucial details in an attempt to meet an assigned compression budget, resulting in a loss of fidelity.
While CoLoR~\citep{CoLoR} also employs a synthetic data method for training a compressor, its data synthesis process, supervision strategy, pipeline design, and target context length are all substantially different from ours.
Despite these efforts, our work addresses the underexplored challenge of compressing significantly longer contexts while offering flexible, user-controllable compression, advancing beyond prior methods limited to shorter inputs and rigid budgets.

%% file: text/3_method.tex
\section{\MODELNAME{}}

\begin{figure*}[t]
  \centering
  \includegraphics[width=0.98\textwidth]{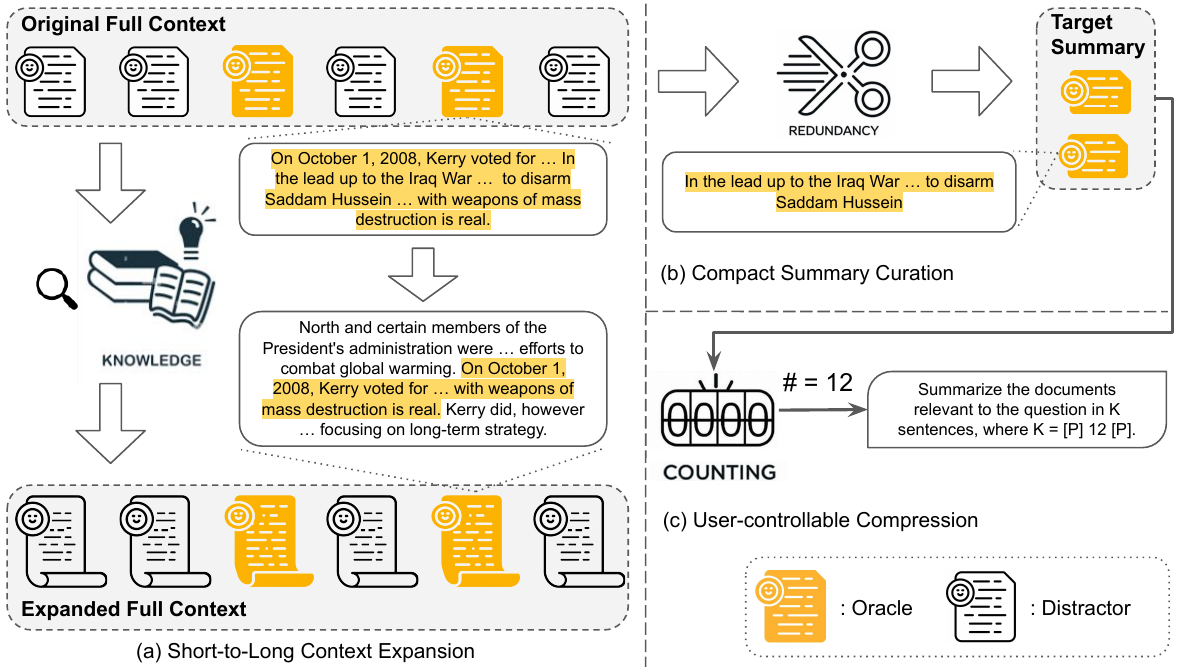}
  \caption{
  An overview of the synthetic data pipeline for training \MODELNAME{}. 
  Starting with a mixture of oracle and distractor documents for a given query, the pipeline:
  (a) expands each document by looking up an external knowledge corpus, 
  (b) curates a compact summary by further reducing redundancy in the oracle documents, and
  (c) generates a user-controllable compression instruction by counting the number of sentences in the compact summary.
  }
  \label{fig-pipeline}
\end{figure*}

  \subsection{Problem Formulation}
  The proposed architecture consists of two modules: a compressor $\mathcal{C}$ and an LM $\mathcal{M}$. 
  For every input query $\mathbf{x}$, an off-the-shelf passage retriever~\citep{DBLP:conf/emnlp/KarpukhinOMLWEC20,DBLP:journals/tmlr/IzacardCHRBJG22} returns a query-related long context $\mathbf{D}$, which may consist of multiple short documents or a single long document. 
  Then, the compressor $\mathcal{C}$ takes as input the concatenation of query $\mathbf{x}$, query-related documents $\mathbf{D}$, and \emph{an optional user-specified compression instruction} $\mathbf{i}$, and outputs a summary $\mathbf{s}$.
  The summary $\mathbf{s}$ captures the core information with respect to $\mathbf{x}$ with significantly fewer words. 
  Finally, the input query $\mathbf{x}$ and the compressed summary $\mathbf{s}$ are fed into an off-the-shelf LM $\mathcal{M}$. 
  The compressor $\mathcal{C}$ is trained on the corpora we curated in this work, while the LM $\mathcal{M}$ remains frozen and can be any off-the-shelf LM.
  In this work, we train a 3B autoregressive model to serve as an \emph{abstractive} compressor $\mathcal{C}$ and adopt a wide range of 8B, 70B, and proprietary models as the LM $\mathcal{M}$. 
  The compressor $\mathcal{C}$ is intentionally designed to be substantially smaller than the LM $\mathcal{M}$, as we aim to reduce the computational cost of encoding the lengthy query-related documents.

  \subsection{\MODELNAME{} Inference}
  Compared to previous work, we propose a novel context compression paradigm featuring both \emph{user-controllable} and \emph{automated} budget allocation.
  For user-controllable compression, we explicitly append a user-specified compression instruction $\mathbf{i}$ that indicates the expected number of sentences in the summary after the context. 
  This direct instruction guides the model to produce a summary of the specified length. 
  In contrast, for automated context compression, the model implicitly determines the optimal number of sentences for the compressed summary based on the inherent content and its internal learned representations.  
  This allows for a more hands-off approach, where the system intelligently decides the summary length, making it ideal for scenarios where a pre-defined sentence count is not feasible or desired.    
  Appendix~\ref{sec-appendix-prompts} presents the detailed inference prompts for the compressor and reader models, respectively.

  \subsection{Data Collection} \label{sec-data}
  Traditional context compression methods can only handle contexts of limited length and often lack flexibility~\citep{DBLP:conf/iclr/XuSC24,DBLP:conf/emnlp/YoonLHJK24,DBLP:conf/naacl/LiGWCP25}, offering limited control over the semantic granularity of the preserved information. 
  In contrast, to develop a more capable and user-controllable context compression model, it is essential to curate a high-quality dataset that captures the nuances of information relevance across various query types. 
  This can help the model learn to prioritize essential context as instructed by the user while eliminating redundancy. 
  This work presents a cost-efficient training recipe by designing a synthetic data pipeline that leverages short-context seed data for long-context compression learning, as illustrated in Figure~\ref{fig-pipeline}. 
  Specifically, our investigation delves into three key practices in terms of expanding the input context, curating the target summary, and designing a user-controllable compression mechanism.

    \subsubsection{Short-to-Long Context Expansion}
    Building models capable of compressing very long contexts requires genuinely long, coherent training examples. Since most available data consists of shorter documents~\citep{DBLP:conf/emnlp/Yang0ZBCSM18,DBLP:journals/tacl/TrivediBKS22}, our goal is to expand these into extended-context data for robust training.

    \paragraph{Source Location}
    For each document, we first identify its topic and use this information to locate the corresponding Wikipedia page from which the document originates. 
    Specifically, the structured Wikipedia corpus released by~\citet{DBLP:journals/tmlr/IzacardCHRBJG22} is leveraged to efficiently retrieve the most relevant page. 
    For documents that cannot be directly matched to an existing Wikipedia page in this corpus, we further turn to the Wikipedia website to search for content on the same topic\footnote{Querying other verifiable, authoritative web sources holds promise for further expanding the scope and ensuring the availability of information beyond fixed, Wikipedia-style text.}.

    \paragraph{Context Expansion}
    Once the appropriate Wikipedia source is located, the precise position of the document within that page is pinpointed.
    Using this position as a reference, the document can be expanded by including a specified number of sentences before and after its original location, thereby naturally enriching the context and extending the content. 
    To control the extent of this expansion, each document is assigned a specific expansion ratio, defined as the number of sentences in the expanded document relative to the original. 
    To diversify context lengths in the training data, the expansion ratio for each document is randomly sampled from a predefined normal distribution with a mean of 20.
    This approach ensures that the expanded context lengths are, on average, substantially longer, while also maintaining a broad range of context lengths to better support model generalization.

    \subsubsection{Compact Summary Curation} \label{sec-summary-curate}
    Given a set of retrieved documents, a pressing research challenge is to identify which text segments within them are the most helpful and can effectively support answering a specific question. 
    Existing QA datasets offer oracle document annotations~\citep{DBLP:conf/emnlp/Yang0ZBCSM18,DBLP:journals/tacl/TrivediBKS22}, and our pipeline operates under the assumption of their availability.
    However, these annotations often exhibit redundancy, which requires additional processing to distill truly essential information.    
    Therefore, to improve the compression rate and extract more precise context, our work extends beyond existing annotations by actively pruning the oracle documents to eliminate this redundancy.

    \paragraph{Helpfulness Definition}
    For a question $\mathbf{x}$, the helpfulness of a sentence $\mathbf{p}_{ij}$ in an oracle document $\mathbf{d}_i$ is determined by the LM's end-task performance when that sentence is removed.
    Formally, we compare the log likelihood assigned to the target output $\mathbf{y}$ by an LM $\mathcal{M}$ before removing the sentence and after. 
    A sentence is considered unhelpful if its removal increases the likelihood of correctly answering that question.

    \paragraph{Head-Tail Iterative Pruning}
    In this work, we explore a practical strategy to achieve this by pruning the head and tail sentences of each oracle document, hypothesizing that critical information is often centrally located.
    For each oracle document, we iteratively check whether each head sentence is unhelpful and remove, continuing this process until a helpful sentence is identified.
    Similarly, we apply the same iterative pruning process to the tail sentences of the document.
    This approach allows us to identify a more compact, continuous, and helpful text segment within the oracle document. 
    Finally, the concatenation of these pruned oracle documents is considered as the target summary.

    \subsubsection{User-controllable Compression}
    
    \paragraph{User-specified Compression Instruction}
    User-controllability is implemented by strategically inserting a clear instruction $\mathbf{i}$ immediately following the provided context. 
    This instruction, phrased as "\emph{Summarize the documents relevant to the question in K sentences, where K = [P] \#\# [\textbackslash P]}" where \#\# is an integer placeholder, empowers the user to directly specify the desired length of the compressed output. 
    By allowing users to assign the number of sentences, we provide a flexible and intuitive mechanism to control the granularity of the summary, ensuring the resulting output aligns precisely with their information needs. 

    \paragraph{Instruction Data Creation}
    To create an instruction-tuning data pair containing: 1) an instruction specifying the number of sentences, and 2) its corresponding summary, we leverage the already constructed summaries in Section~\ref{sec-summary-curate}. 
    For each summary, we count the number of sentences it contains, let's say this count is \emph{k}. 
    Then, we formulate the instruction part of the pair as "\emph{Summarize the documents relevant to the question in K sentences, where K = [P] k [\textbackslash P]}."
    The corresponding summary for this instruction is simply the summary we initially constructed. 
    This direct pairing ensures that the model learns the precise relationship between a numerical sentence constraint in the instruction and the actual length of the summary. 
    By generating a diverse set of such pairs across various contexts and their pre-computed summaries, we build a robust dataset for instruction tuning, enabling the model to generalize and produce summaries of user-specified lengths.

  \subsection{\MODELNAME{} Training}
  The curated dataset $\mathcal{D}_{comp}$ is utilized to fine-tune the compressor $\mathcal{C}$, a \texttt{Llama-3.2-3B-Instruct} model~\citep{meta2024llama3_2_release}. 
  This model strikes an effective balance between computational efficiency and performance. Its relatively compact size, compared to larger LLMs, makes it well-suited for context compression without demanding excessive computational resources.
  The fine-tuning process follows the standard next token objective, formulated as:
  \begin{equation}
    \max_{\mathcal{C}}\mathbb{E}_{(\mathbf{x}, \mathbf{D}, \mathbf{i}, \mathbf{s}) \sim \mathcal{D}_{comp}} \log p_{\mathcal{C}} (\mathbf{s} | \mathbf{x}, \mathbf{D}, \mathbf{i}). 
    \label{equ-objective}
  \end{equation}

  \paragraph{Discussion}
   We respectfully note that the novelty of our method lies not merely in combining existing techniques but in the careful design and training of a lightweight compression model that achieves strong long-context summarization at low inference cost, preserves downstream task quality, and generalizes across diverse datasets. Achieving this balance is non-trivial and requires tailored training strategies, aspect-based supervision, and length-controlled outputs, which are not provided by off-the-shelf summarization or LLM-based pipelines. 

%% file: text/4_experiment.tex
\section{Experiments}

  \subsection{Experimental Settings}

\input{table/tab_train_data_stats}

    \paragraph{\MODELNAME{} Series}
    \MODELNAME{} was initialized from \texttt{Llama-3.2-3B-Instruct}~\citep{meta2024llama3_2_release}. 
    The original training sets of MuSiQue~\citep{DBLP:journals/tacl/TrivediBKS22}, HotpotQA~\citep{DBLP:conf/emnlp/Yang0ZBCSM18}, and LongAlign~\citep{DBLP:conf/emnlp/BaiLZHQH0DL24} were used as the seed data of our synthesis pipeline.
    Table~\ref{tab-train-data-stats} presents the statistics of the curated training data. 
    We evaluated a series of \MODELNAME{} models that feature diverse user-controllable and automated compression scenarios.
    In the user-controllable setting, we defined three compression levels of \textsc{High}, \textsc{Medium}, and \textsc{Low}, which were empirically set to compress to 5, 10, and 20 sentences, respectively.
    The \textsc{Auto} mode denotes dynamic thinking, where the model decides the number of sentences based on the complexity of the task.

    \paragraph{Reader Series}
    To demonstrate that \MODELNAME{} can benefit a wide range of models, small (\texttt{Llama-3.1-8B-Instruct}), large (\texttt{Llama-3.1-70B-Instruct}), and proprietary (\texttt{GPT-4.1-nano})\footnote{\texttt{gpt-4.1-nano-2025-04-14} for its favorable cost.} language models were used as the reader $\mathcal{M}$. 
    Readers may refer to Appendix~\ref{sec-implement} for more implementation details.

\input{table/tab_test_data_stats}

\input{table/tab_main_v2}

    \paragraph{Datasets}
    We evaluated the \MODELNAME{} series on the following four open-domain multi-hop QA datasets: MuSiQue~\citep{DBLP:journals/tacl/TrivediBKS22}, HotpotQA~\citep{DBLP:conf/emnlp/Yang0ZBCSM18}, and 2WikiMultiHopQA~\citep{DBLP:conf/coling/HoNSA20} which are extended versions from LongBench~\citep{DBLP:conf/acl/BaiLZL0HDLZHDTL24}, and LongSeal~\citep{DBLP:journals/corr/abs-2506-01062}.
    Table~\ref{tab-test-data-stats} presents the statistics of the evaluation data.
    Additionally, to further demonstrate the superior cross-domain robustness of \MODELNAME{} over previous compression methods, we refer readers to Table~\ref{tab:main-results-OOD} in Appendix~\ref{sec-results} for results on three non-Wikipedia-based datasets.

    \paragraph{Metrics}
    \textbf{Exact match (EM)} and \textbf{F1} of answer strings were reported for QA performance.
    \textbf{Compression rate} was defined as the ratio of the number of words in the retrieved documents before compression to the number of words in the compressed summary after compression.
    A higher compression rate indicates a shorter summary.

    \paragraph{Baselines}
    We compared the \MODELNAME{} series with four main categories.
    (1) \textbf{Long-context LLMs} including: 
    \textbullet{} \underline{FILM-7B}~\citep{DBLP:conf/nips/AnML0LC24}. 
    \textbullet{} \underline{ProLong-8B}~\citep{gao2025how}. 
    They present training recipes to enhance the long-context capabilities. 
    (2) \textbf{Non-compression} denotes prepending the full retrieved documents without compression.
    (3) \textbf{Extractive compression} methods including: 
    \textbullet{} \underline{RECOMP (Extractive)}~\citep{DBLP:conf/iclr/XuSC24} ranks sentences based on whether it is useful as input for LM.
    \textbullet{} \underline{EXIT}~\citep{DBLP:conf/acl/HwangCJSHP25} classifies sentence-level relevance with lightweight single-token predictions, and reassembles only the high-relevance sentences in their original order. 
    \textbullet{} \underline{Rerank Top-\emph{k}} denotes reranking the set of retrieved documents and keeping only the top-\emph{k} documents. 
    (4) \textbf{Abstractive compression} methods including: 
    \textbullet{} \underline{RECOMP (Abstractive)}~\citep{DBLP:conf/iclr/XuSC24} distills the summarization knowledge of proprietary LLMs (\texttt{gpt-3.5-turbo}) into an abstractive compressor T5-large.
    \textbullet{} \underline{BRIEF}~\citep{DBLP:conf/naacl/LiGWCP25} enhances compression necessitating multi-hop reasoning by curating synthetic data. 
    \textbullet{} \underline{Llama-3.1-3B-Instruct} denotes the off-the-shelf official release without further fine-tuning.
    \textbullet{} \underline{LongLLMLingua}~\citep{DBLP:conf/acl/JiangWL0L0Q24}  performs both demonstration-level and token-level compression, leveraging their perplexity calculated by a causal LM \texttt{Llama-2-7B-Chat}. 
    \textbullet{} \underline{GPT-4.1-nano} was prompted to summarize the documents with respect to the question. 
    \textbullet{} \MODELNAME{}-\textsc{Auto}$_{\text{L7C}}$ denotes that \MODELNAME{}-\textsc{Auto} is initialized from \underline{L}lama2-\underline{7}B-\underline{C}hat, allowing a fair head-to-head comparison with LongLLMLingua under the same base model. Note that the context length of Llama-2-7B-Chat is 4K tokens, which is shorter than the average length of our training data. Consequently, we truncate the distractor documents in our training data, which substantially degrades the performance of \MODELNAME{}-\textsc{Auto}$_{\text{L7C}}$ relative to \MODELNAME{}-\textsc{Auto}, while still outperforming LongLLMLingua.
    Readers may refer to Appendix~\ref{sec-baselines} for more baseline details.

  \subsection{Experimental Results}
  Table~\ref{tab:main-results-v2} presents the evaluation results on four multi-hop QA tasks with three different reader models.
  \MODELNAME{} demonstrates promising multi-hop performance in both QA and document compression.
  Compared to non-compression, \MODELNAME{}-\textsc{Auto} achieves an average compression rate of 32x, while still outperforming it by 6.70\%, 0.60\%, and 7.27\% across three different reader models, respectively. These results indicate that existing LLMs still struggle with the demands of heavy long-context understanding, while a lightweight, highly specialized long-context compressor can help identify relevant information and alleviate the cognitive burden on reader models, enabling fast and accurate multi-hop reasoning.
  Although the curated training data for FILM-7B and ProLong-8B improves their long-context capabilities, their performance still underperforms \MODELNAME{} significantly while also requiring more computational resources.
  Compared to LongLLMLingua, \MODELNAME{}-\textsc{Auto} compresses by higher 32x than its 9x on average, while still outperforming it by 6.77\%, 4.67\%, and 7.77\%, respectively.
  Compared to using the proprietary \texttt{GPT-4.1-nano} as the compressor, it tends to significantly compress long contexts which results in suboptimal QA performance.
  For the \MODELNAME{} series, the three compression levels offer varying granularities in preserving key semantic information, as guided by the user-specified instruction. This balance of efficiency and accuracy demonstrates its robustness and versatility across diverse scenarios.

  \subsection{Analysis}

    \begin{figure}[t]
      \centering
      \includegraphics[width=0.45\textwidth]{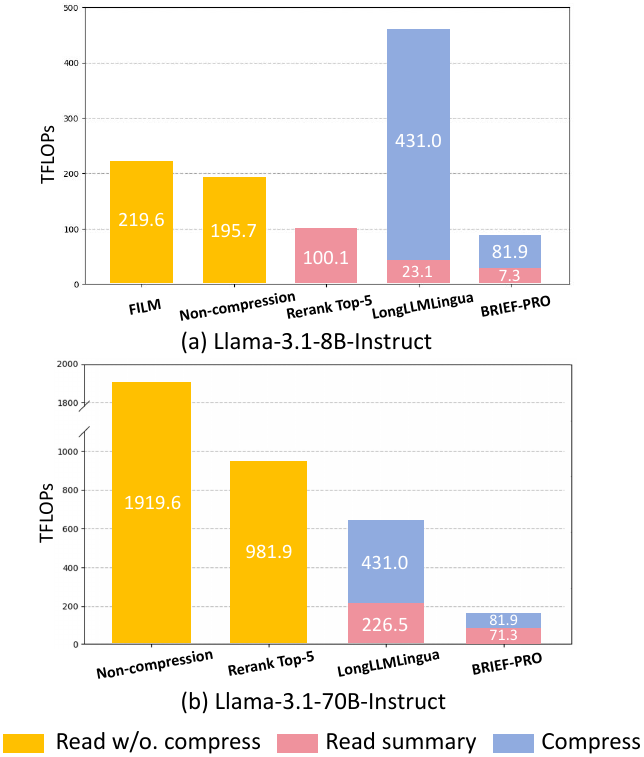}
      \caption{The comparison of TFLOPs consumption using (a) \texttt{Llama-3.1-8B-Instruct} and (b) \texttt{Llama-3.1-70B-Instruct} as the reader model.}
      \label{fig-analysis-flops}
    \end{figure}

    \paragraph{The improvement of latency in terms of overall computational overhead}
    Figure~\ref{fig-analysis-flops} presents a comparison of TFLOPs consumption for processing long contexts. 
    The profiler provided by Accelerate to count flops was adopted\footnote{\url{https://huggingface.co/docs/accelerate/usage\_guides/profiler}}.
    Specifically, when \MODELNAME{} is adopted for compression, the overall required TFLOPs are significantly lower than those needed for the original, uncompressed long contexts.
    The total amount of computation is reduced to 45\% and 8\% of what it was before compression using \texttt{Llama-3.1-8B-Instruct} and \texttt{Llama-3.1-70B-Instruct} as the LM $\mathcal{M}$, respectively. 
    Compared to LongLLMLingual which uses \texttt{Llama-2-7B-Chat} to compute sentence and token perplexity, \MODELNAME{} consumes less than 20\% and 24\% of LongLLMLingual’s resources, respectively, while delivering better performance.
    LongLLMLingual incurs higher computational costs with the 8B reader model, because it has to divide intermediate results into segments and apply token-level compression iteratively, where each token’s perplexity based on preceding compressed segments.
    This substantial reduction in TFLOPs highlights \MODELNAME{}'s potential to optimize inference, especially for large-scale long context and with larger reader models, by enabling reader models to focus on compressed, more relevant information without sacrificing accuracy.
    We also analyzed end-to-end latency by summing the execution times of all pipeline components. Notably, with the 70B model, the overall latency was reduced to 14\%, 32\%, and 7\% of that of Non-Compression, Rerank Top-5, and LongLLMLingua, respectively (see Appendix~\ref{sec-results} for details).

\input{table/tab_expand_oracle_only_avg}

    \paragraph{The comparison with exhaustively expanding only oracle documents}
    In real-world scenarios, detailed knowledge about a fact may already be available. This raises the question: what would be the impact of \emph{directly using this ready-made knowledge} to construct a long context?
    To demonstrate the effectiveness of expanding both oracle and distractor documents to form the input long context, our approach is compared against a strategy that performs exhaustive expansion solely on oracle documents, utilizing complete Wikipedia pages. 
    The experimental results shown in Table~\ref{tab-expand-oracle-only-avg} demonstrate a significant performance degradation when only oracle documents are expanded. 
    This suggests that expanding only oracle documents might lead to a somewhat artificially "clean" context, potentially overestimating the model's ability to handle complex, noisy inputs.
    In contrast, incorporating the expansion of distractor documents provides essential contextual diversity and better reflects realistic long-context scenarios, where relevant and irrelevant information is often interspersed.
    On the other hand, our method is statistically able to synthesize significantly longer contexts (Avg. 6.0k vs. 3.6k words).
    The ability to process and learn from these substantially longer, noisy contexts directly contributes to the observed performance gains.

    \input{table/tab_control_error}

    \paragraph{The accuracy of user-controllable instructions in terms of target sentence count}
    The accuracy of the instruction hinges on how precisely the system can adhere to the user-specified sentence count. 
    While the intent is to provide a flexible and intuitive mechanism for controlling summary granularity, the actual accuracy of this control can vary. 
    Table~\ref{tab-control-error} presents the average sentence counts in the generated summary in various compression modes across all four test sets. 
    Appendix~\ref{sec-results} presents the detailed distribution.
    Although fitting the summary perfectly within the specified sentence limit is challenging, the results show that \MODELNAME{} performs well in following the \textsc{High} and \textsc{Medium} compression instructions. 
    The reason lies in the sufficient training data for the target summaries within this length range.
    Meanwhile, achieving precise control while maintaining high summarization quality remains a technical challenge and should continue to be an active area of research.

%% file: table/tab_train_data_stats.tex
\begin{table}[t]
  \centering
  \setlength{\tabcolsep}{3.0pt}
  \begin{tabular}{p{4cm} >{\centering\arraybackslash}p{2cm}}
    \toprule

    \textbf{Training Set Statistic} & \textbf{Number}  \\

    \midrule
    \textbf{Sample Number}           &    45.2k      \\
    \textbf{Context Words}           &               \\
    - Average                        &    6.0k       \\
    - Standard Deviation             &    3.5k        \\
    \textbf{Summary Words}           &               \\
    - Average                        &    0.2k     \\
    - Standard Deviation             &    0.3k       \\
    
    \bottomrule
  \end{tabular}  
  \caption{The statistics of the curated training data.}
  \label{tab-train-data-stats}
\end{table}

%% file: table/tab_test_data_stats.tex
\begin{table}[t]
  \centering
  \setlength{\tabcolsep}{3.0pt}
  \begin{tabular}{lcc}
  \toprule
                     & \# samples & \# context words \\
  \midrule
    MuSiQue          &    200     & 11.2k  \\
    HotpotQA         &    200     & 9.2k   \\
    2WikiMultiHopQA  &    200     & 4.9k   \\
    LongSeal         &    254     & 14.8k  \\
  \bottomrule
  \end{tabular}
  \caption{The statistics of the evaluation data.}
  \label{tab-test-data-stats}
\end{table}

%% file: table/tab_main_v2.tex
\begin{table*}[ht!]
\centering
\resizebox{\linewidth}{!}{
\def\arraystretch{1.1}
\begin{tabular}{llcccccccccccccc}
\toprule
  \multirow{2}{*}{\textbf{Category}} & \multirow{2}{*}{\textbf{Method}} &  \multicolumn{3}{c}{\textbf{MuSiqQue}}   & \multicolumn{3}{c}{\textbf{HotpotQA}} & \multicolumn{3}{c}{\textbf{2WikiMultiHopQA}} & \multicolumn{3}{c}{\textbf{LongSeal}} & \multicolumn{2}{c}{\textbf{Average}} \\
                                   & & \textbf{EM} & \textbf{F1} & \textbf{Rate} & \textbf{EM} & \textbf{F1} & \textbf{Rate} & \textbf{EM} & \textbf{F1} & \textbf{Rate} & \textbf{EM} & \textbf{F1} & \textbf{Rate} & \textbf{QA} & \textbf{Rate} \\
\midrule
 \multirow{2}{*}{Long-context LLMs} & ProLong-8B & 19.50 & 27.42 & 1x & 42.50 & 54.43 & 1x & 36.00 & 42.94 & 1x & 7.09 & 11.71 & 1x & 30.20 & 1x \\
                                    & FILM-7B & 26.50 & 36.45 & 1x & 47.50 & 61.43 & 1x & 38.50 & 46.14 & 1x & 9.84 & 15.10 & 1x & 35.18 & 1x\\
\midrule
\rowcolor{gray!20} \multicolumn{16}{c}{\textbf{\texttt{Llama-3.1-8B-Instruct}}} \\
\midrule
  \multirow{1}{*}{Non-compression} & & 20.50 & 28.88 & 1x & 40.00 & 54.29 & 1x & 37.00 & 46.63 & 1x & 12.60 & 16.81 & 1x & 32.09 & 1x  \\
\cdashline{1-16}
 \multirow{5}{*}{Extractive}
 
                          & RECOMP (Extractive) &  16.00  &  23.39  &  41x &  35.00  &  48.48  &  36x  & 27.50 & 35.40 &  20x &  4.72  &  8.18  & 35x &  24.83  &  33x \\

                          & EXIT &  11.00  &  17.19  &  35x  &  36.50  &  48.62  &  32x  &  36.00  &  41.83  &  16x  &  7.09 &  10.22  &  20x  &  26.06  &  26x  \\

                          & Rerank Top-1 & 9.50 & 14.58 & 7x & 23.50 & 35.66 & 8x & 13.00 & 20.75 & 11x & 6.30 & 9.08 & 15x & 16.55 & 10x \\
                          
                          & Rerank Top-3 & 15.50 & 23.40 & 2.5x & 37.50 & 50.07 & 2.2x & 36.50 & 46.14 & 3.0x & 7.87 & 12.19 & 4.2x & 28.65 & 3x \\

                          & Rerank Top-5 & 20.50 & 28.03 & 1.7x & 42.00 & 55.16 & 1.5x & 43.00 & 52.74 & 1.8x & 9.06 & 12.42 & 2.7x & 32.86 & 2x \\
                          
\cdashline{1-16}
 \multirow{10}{*}{Abstractive}& BRIEF & 5.00 & 12.39 & 22x & 18.00 & 27.70 & 59x & 21.00 & 26.38 & 28x & 5.91 & 10.02 & 19x & 15.80 & 32x \\
 
                              & RECOMP (Abstractive) &  17.50  &  25.92  &  10x  &  28.50  &  40.59  &  10x  &  19.50  &  26.60  &  9x  &  5.51  &  9.88  &  9x  &  21.75  &  10x \\

                              & Llama-3.2-3B-Instruct & 13.00 & 18.48 & 48x & 30.00 & 41.67 & 41x & 34.50 & 43.29 & 39x & 7.48 & 12.03 & 58x & 25.06 & 47x \\
 
                              & LongLLMLingua & 20.50 & 28.75 & 10x & 38.50 & 53.63 & 8x & 43.00 & 50.24 & 5x & 8.27 & 13.26 & 14x & 32.02 & 9x \\
                              
                              & GPT-4.1-nano & \textbf{28.50} & \textbf{40.33} & 128x & 38.00 & 53.67 & 90x & \textbf{47.50} & \textbf{60.34} & 108x & \underline{10.63} & \underline{15.95} & 116x & \underline{36.87} & 110x \\

                              & \MODELNAME{}-\textsc{Auto}$_{\text{L7C}}$ &  22.50  &  32.44  &  13x  &  \underline{42.50}  &  \underline{56.17}  &  13x  &  43.00  &  51.55  &  14x  &  6.12  &  10.59  &  11x  &  33.11  &  12x  \\
                              
                              & \MODELNAME{}-\textsc{Auto} & \underline{27.50} & \underline{36.64} & 35x & \textbf{49.00} & \textbf{63.05} & 34x & \textbf{47.50} & \underline{56.00} & 23x & \textbf{12.99} & \textbf{17.62} & 36x & \textbf{38.79} & 32x\\

                              \cdashline{2-16}

                              & \MODELNAME{}-\textsc{High} & 24.00 & 31.70 & 84x & 47.00 & 58.85 & 72x & 40.00 & 47.79 & 45x & 9.06 & 13.73 & 72x & 34.02 & 68x\\

                              & \MODELNAME{}-\textsc{Medium} & 27.50 & 35.55 & 47x & 51.00 & 63.41 & 42x & 48.50 & 57.21 & 26x & 10.24 & 15.67 & 52x & 38.64 & 42x\\
 
                              & \MODELNAME{}-\textsc{Low} & 30.50 & 39.68 & 26x & 50.50 & 64.36 & 26x & 49.00 & 58.39 & 15x & 11.42 & 16.63 & 31x & 40.06 & 25x\\

 \midrule
\rowcolor{gray!20} \multicolumn{16}{c}{\textbf{\texttt{Llama-3.1-70B-Instruct}}} \\
\midrule
  \multirow{1}{*}{Non-compression} & & 36.00 & 45.47 & 1x & 51.50 & 65.14 & 1x & 59.50 & 66.85 & 1x & 15.35 & 20.05 & 1x & 44.98 & 1x\\
\cdashline{1-16}
 \multirow{5}{*}{Extractive}

                          & RECOMP (Extractive) &  29.50  &  37.65  &  41x  &  40.00  &  53.27  &  36x  &  40.50  &  47.46  &  20x  &  5.91  &  9.43  &  35x  &  32.97  &  33x \\

                          & EXIT &  23.00  &  30.85  &  35x  &  46.00  &  59.80  &  32x  &  52.00  &  58.93  &  16x  &  7.87  & 11.19  &  20x  &  36.21  &  26x  \\

                          & Rerank Top-1 & 19.00 & 26.50 & 7x & 32.00 & 44.42 & 8x & 21.00 & 27.14 & 11x & 6.69 & 9.23 & 15x & 23.25 & 10x \\
                          
                          & Rerank Top-3 & 26.00 & 35.32 & 2.5x & 43.00 & 55.78 & 2.2x & 49.50 & 58.75 & 3.0x & 10.63 & 13.21 & 4.2x & 36.52 & 3x  \\

                          & Rerank Top-5 & 30.00 & 39.06 & 1.7x & 50.00 & 64.45 & 1.5x & 57.50 & 65.78 & 1.8x & 11.02 & 14.57 & 2.7x & 41.55 & 2x \\

\cdashline{1-16}
 \multirow{10}{*}{Abstractive}& BRIEF & 11.50 & 18.74 & 22x & 25.00 & 36.03 & 59x & 31.00 & 36.53 & 28x & 7.09 & 11.13 & 19x & 22.13 & 32x \\
 
                              & RECOMP (Abstractive) &  25.00 &  33.80  &  10x  &  37.50  &  50.47  &  10x  &  34.50  &  41.56  &  9x  &  7.09  &  11.10  &  9x  &  30.13  &  9.5x \\

                              & Llama-3.2-3B-Instruct & 24.50 & 30.66 & 48x & 34.50 & 47.26 & 41x & 42.00 & 50.12 & 39x & 9.45 & 12.48 & 58x & 31.37 & 47x \\

                              & GPT-4.1-nano & 32.00 & 41.46 & 128x & 42.00 & 56.59 & 90x & 51.00 & 64.03 & 108x & \underline{10.63} & \underline{16.55} & 116x & 39.28 & 110x \\
 
                              & LongLLMLingua & 32.00 & 42.35 & 10x & 46.00 & \underline{61.32} & 8x & \underline{55.50} & \underline{64.96} & 5x & 10.24 & 14.87 & 14x & 40.91 & 9x \\

                              & \MODELNAME{}-\textsc{Auto}$_{\text{L7C}}$ &  \underline{36.00}  &  \underline{46.40}  &  13x  &  \underline{47.50}  &  60.97  &  13x  &  54.50  &  62.68  &  14x  &  9.06  &  12.68  &  11x  &  \underline{41.22}  &  12x  \\
                              
                              & \MODELNAME{}-\textsc{Auto} & \textbf{38.50} & \textbf{48.84} & 35x & \textbf{53.00} & \textbf{67.29} & 34x & \textbf{59.50} & \textbf{66.94} & 23x & \textbf{12.99} & \textbf{17.59} & 36x & \textbf{45.58} & 32x\\

                              \cdashline{2-16}
                              
                              & \MODELNAME{}-\textsc{High} & 38.00 & 45.55 & 84x & 55.00 & 67.42 & 72x & 54.00 & 60.50 & 45x & 9.45 & 14.18 & 72x & 43.01 & 68x\\

                              & \MODELNAME{}-\textsc{Medium} & 38.00 & 49.17 & 47x & 54.50 & 68.02 & 42x & 58.50 & 67.38 & 26x & 10.63 & 16.41 & 52x & 45.33 & 42x\\
 
                              & \MODELNAME{}-\textsc{Low} & 40.00 & 51.95 & 26x & 54.50 & 68.35 & 26x & 59.50 & 68.31 & 15x & 12.60 & 16.72 & 31x & 46.49 & 25x\\
                              
\midrule
\rowcolor{gray!20} \multicolumn{16}{c}{\textbf{\texttt{GPT-4.1-nano}}} \\
\midrule
  \multirow{1}{*}{Non-compression} & & 24.00 & 33.08 & 1x & 42.50 & 56.56 & 1x & 39.00 & 46.04 & 1x & 11.02 & 16.02 & 1x & 33.53 & 1x \\
\cdashline{1-16}
 \multirow{5}{*}{Extractive}  
 
                          & RECOMP (Extractive) &  19.50  &  27.32  &  41x  &  41.00  &  52.44  &  36x  &  38.00  &  44.81  &  20x  &  3.54  &  7.06  &  35x  &  29.21  &  33x \\

                          & EXIT &  16.50  &  24.33  &  35x  &  42.00  &  54.41  &  32x  &  43.00  &  47.58  &  16x  &  6.69  &  10.65  &  20x  &  30.65  &  26x  \\

                          & Rerank Top-1 & 11.00 & 17.99 & 7x & 30.00 & 40.85 & 8x & 23.50 & 27.73 & 11x & 3.94 & 6.55 & 15x & 20.20 & 10x  \\

                          & Rerank Top-3 & 22.50 & 30.73 & 2.5x & 40.00 & 52.85 & 2.2x & 45.50 & 52.83 & 3.0x & 7.87 & 11.83 & 4.2x & 33.01 & 3x \\
                          
                          & Rerank Top-5 & 21.00 & 29.02 & 1.7x & 41.00 & 54.87 & 1.5x & 46.00 & 54.71 & 1.8x & 9.45 & 13.99 & 2.7x & 33.76 & 2x \\

\cdashline{1-16}
 \multirow{10}{*}{Abstractive}& RECOMP (Abstractive) &  13.00  &  19.13  &  10x  &  27.00 &  39.21  &  10x  &  25.50  &  30.26  &  9x  &  3.54 &  8.11  &  9x  &  20.72  &  9.5x \\
 
                              & BRIEF & 6.00 & 13.55 & 22x & 25.50 & 34.75 & 59x & 30.00 & 35.92 & 28x & 8.27 & 12.30 & 19x & 20.79 & 32x \\

                              & Llama-3.2-3B-Instruct & 17.50 & 23.60 & 48x & 33.00 & 44.72 & 41x & 38.50 & 47.10 & 39x & 8.27 & 11.76 & 58x & 28.06 & 47x \\
 
                              & LongLLMLingua & 20.50 & 28.55 & 10x & 40.50 & 54.90 & 8x & 48.00 & 54.52 & 5x & 5.91 & 11.35 & 14x & 33.03 & 9x  \\
                              
                              & GPT-4.1-nano & \textbf{31.50} & \textbf{41.24} & 128x & 40.50 & 55.87 & 90x & \textbf{51.50} & \textbf{63.79} & 108x & \underline{11.02} & \underline{16.51} & 116x & \underline{38.99} & 110x \\

                              & \MODELNAME{}-\textsc{Auto}$_{\text{L7C}}$ &  20.50  &  31.58  &  13x  &  \underline{46.00}  &  \underline{56.06}  &  13x  &  46.50  &  52.81  &  14x  &  7.87  &  11.35  &  11x  &  34.08  &  12x  \\
                              
                              & \MODELNAME{}-\textsc{Auto} & \underline{29.50} & \underline{41.10} & 35x & \textbf{51.50} & \textbf{65.52} & 34x & \underline{51.00} & \underline{58.68} & 23x & \textbf{12.20} & \textbf{16.87} & 36x & \textbf{40.80} & 32x \\

                              \cdashline{2-16}

                              & \MODELNAME{}-\textsc{High} & 31.50 & 38.70 & 84x & 51.00 & 64.13 & 72x & 49.00 & 56.48 & 45x & 11.02 & 14.91 & 72x & 39.59 & 68x\\

                              & \MODELNAME{}-\textsc{Medium} & 30.50 & 41.15 & 47x & 52.00 & 65.38 & 42x & 53.50 & 62.08 & 26x & 11.02 & 15.31 & 52x & 41.37 & 42x\\
 
                              & \MODELNAME{}-\textsc{Low} & 33.00 & 42.96 & 26x & 51.00 & 64.86 & 26x & 52.00 & 60.22 & 15x & 11.42 & 15.76 & 31x & 41.40 & 25x\\
                     
\bottomrule
\end{tabular}
}
\caption{Evaluation results on four multi-hop QA tasks with small, large, and proprietary LMs as the $\mathcal{M}$, respectively. 
\textbf{Bold} and \underline{underscore} denote the best and second-best QA performance, respectively, under the model’s self-determined compression setting (i.e., the \textsc{Auto} mode).
}

\label{tab:main-results-v2}
\end{table*}

%% file: table/tab_expand_oracle_only_avg.tex
\setlength{\tabcolsep}{2.5pt}
\begin{table}[t]
\centering
\begin{tabular}{lccc}
\toprule
  \multirow{2}{*}{\textbf{Training Data}} & \multirow{2}{*}{\textbf{Length}} & \multicolumn{2}{c}{\textbf{Average}} \\
                                   & & \textbf{QA} & \textbf{Rate} \\
\midrule

\rowcolor{gray!20} \multicolumn{4}{c}{\textbf{\texttt{Llama-3.1-8B-Instruct}}} \\
\midrule
  Oracle++ \& Distractor++ & 6.0k & 38.79 & 32x\\
  Oracle+ \& Distractor+   & 3.6k & 36.02 & 34x\\
  Oracle+++                & 3.6k & 33.76 & 35x\\
 \midrule
 
\rowcolor{gray!20} \multicolumn{4}{c}{\textbf{\texttt{Llama-3.1-70B-Instruct}}} \\
\midrule              
  Oracle++ \& Distractor++ & 6.0k & 45.58 & 32x\\
  Oracle+ \& Distractor+   & 3.6k & 41.74 & 34x\\
  Oracle+++                & 3.6k & 41.68 & 35x\\                       
\midrule

\rowcolor{gray!20} \multicolumn{4}{c}{\textbf{\texttt{GPT-4.1-nano}}} \\
\midrule                   
  Oracle++ \& Distractor++ & 6.0k & 40.80 & 32x \\
  Oracle+ \& Distractor+   & 3.6k & 39.11 & 34x\\
  Oracle+++                & 3.6k & 37.03 & 35x \\                     
\bottomrule
\end{tabular}
\caption{Evaluation results, averaged over four test sets, of comparing with exhaustively expanding only oracle documents. 
The number of + denotes the extent of expansion.
Appendix~\ref{sec-results} presents the full results.}
\label{tab-expand-oracle-only-avg}
\end{table}

%% file: table/tab_control_error.tex
\begin{table}[t]
  \centering
  \setlength{\tabcolsep}{5.0pt}
  \begin{tabular}{ccc}
  \toprule
    Compression Mode & Expected & Average \\
  \midrule
    \textsc{High}    &    5     &   6.2   \\
    \textsc{Medium}  &   10     &  10.4   \\
    \textsc{Low}     &   20     &  18.0   \\
  \bottomrule
  \end{tabular}  
  \caption{Average sentence counts in the generated summary in various compression modes.}
  \vspace{-2mm}
  \label{tab-control-error}
\end{table}

%% file: text/5_conclusion.tex
\section{Conclusion}
This work introduces \MODELNAME{}, a universal, lightweight context compressor tailored for long contexts of 10k+ words across diverse scenarios, enabling fast and accurate multi-hop reasoning with RAG. 
\MODELNAME{} is trained on synthetic data generated through a pipeline built from short-context seed data, which synthesizes long-context training examples for compression learning. 
This pipeline provides a data-centric approach that supports user-controllable compression over output summary length. 
Experimental results demonstrate that the denoised summaries produced by \MODELNAME{} substantially reduce computational overhead while improving QA accuracy across a wide range of small, large, and proprietary language models compared to previous compression methods.

\section*{Limitations}
While \MODELNAME{} demonstrates promising results in long-context compression for RAG, several potential limitations warrant consideration. 
First, the model's ability to effectively abstract and compress information from significantly longer (e.g., more than 20k words) and more complex inputs than those seen during training could be constrained.
It could potentially lead to a performance degradation on contexts vastly exceeding the training data's length or complexity.
Second, although \MODELNAME{} is highly effective within the RAG framework, its performance in other long-context applications, such as few-shot learning, code completion, or long-dialogue history understanding, remains untested and could be suboptimal. 
Take code completion as an example, the task demands strict adherence to syntax, accurate tracking of variable definitions, precise function signatures, and coherent logical structure. 
A compressor optimized for natural language evidence may struggle to maintain the exactness and completeness required for reliable code generation.
Without a comprehensive evaluation across these diverse long-context tasks, the generalizability and efficacy of \MODELNAME{} beyond its RAG-specific domain remain an open question.

%% file: text/_appendix.tex
\section{Prompts} \label{sec-appendix-prompts}

The following are the prompts for the compression model and the reader model. For the reader model, we follow the prompt used in LongBench~\citep{DBLP:conf/acl/BaiLZL0HDLZHDTL24}.

\begin{tcolorbox}[
    colframe=black,
    colback=white, 
    coltitle=white, 
    colbacktitle=black, 
    title=Prompt for Compression Model (Auto),
    fonttitle=\bfseries,
    boxrule=1pt,
    arc=2mm,    
    left=4mm, right=4mm, top=3mm, bottom=3mm
]
{
\setlength{\parskip}{3pt}

Write a high-quality summary of the provided documents with respect to the question.

\#\#\# This is the question: \{QUESTION\}

\#\#\# These are the documents:

\{DOCUMENTS\}

\#\#\# This is the summary:

}

\end{tcolorbox}

\begin{tcolorbox}[
    colframe=black,
    colback=white, 
    coltitle=white, 
    colbacktitle=black, 
    title=Prompt for Compression Model (User-controllable),
    fonttitle=\bfseries,
    boxrule=1pt,
    arc=2mm,    
    left=4mm, right=4mm, top=3mm, bottom=3mm
]
{
\setlength{\parskip}{3pt}

Write a high-quality summary of the provided documents with respect to the question.

\#\#\# This is the question: \{QUESTION\}

\#\#\# These are the documents:

\{DOCUMENTS\}

\#\#\# This is the summary:

Summarize the documents relevant to the question in K sentences, where K = [P] \{LENGTH\} [\textbackslash P]
}

\end{tcolorbox}

\begin{tcolorbox}[
    colframe=black,
    colback=white, 
    coltitle=white, 
    colbacktitle=black, 
    title=Prompt for Reader Model,
    fonttitle=\bfseries,
    boxrule=1pt,
    arc=2mm,    
    left=4mm, right=4mm, top=3mm, bottom=3mm
]
{
\setlength{\parskip}{3pt}

Answer the question based on the given passages. Only give me the answer and do not output any other words.

The following are given passages.

\{DOCUMENTS\}

Answer the question based on the given passages. Only give me the answer and do not output any other words.

Question: \{QUESTION\}

Answer:

}

\end{tcolorbox}

\clearpage

\section{Experimental Details}

  \subsection{Baselines} \label{sec-baselines}
    We compared the \MODELNAME{} series with four main categories:
    
    (1) \textbf{Long-context LLMs} including: 
    \textbullet{} \underline{FILM-7B}~\citep{DBLP:conf/nips/AnML0LC24} initialized from Mistral-7B~\citep{jiang2023mistral}.  
    \textbullet{} \underline{ProLong-8B}~\citep{gao2025how} initialized from Llama-3-8B~\citep{dubey2024llama}. 
    They present training recipes to enhance long-context capabilities. FILM leverages synthesized, information-intensive long-context QA data, while ProLong combines long-context code repositories and books with high-quality short-context data.
    
    (2) \textbf{Non-compression} denotes prepending the full retrieved documents without compression.
    
    (3) \textbf{Extractive compression} methods including: 
    \textbullet{} \underline{RECOMP (Extractive)}~\citep{DBLP:conf/iclr/XuSC24} formulates extractive compression as a sentence ranking problem, and the sentence is evaluated based on whether it is useful as input for the LM.
    \textbullet{} \underline{EXIT}~\citep{DBLP:conf/acl/HwangCJSHP25} splits a document into sentences, classifies sentence-level relevance with lightweight single-token predictions, and reassembles only the high-relevance sentences in their original order to preserve coherence and key information.
    \textbullet{} \underline{Rerank Top-\emph{k}} denotes reranking the set of retrieved documents using Contriever~\citep{DBLP:journals/tmlr/IzacardCHRBJG22} trained on MS MARCO dataset and keeping only the top-\emph{k} documents. 
    
    (4) \textbf{Abstractive compression} methods including: 
    \textbullet{} \underline{RECOMP (Abstractive)}~\citep{DBLP:conf/iclr/XuSC24} distills the summarization knowledge of proprietary LLMs (\texttt{gpt-3.5-turbo}) into an abstractive compressor T5-large.
    \textbullet{} \underline{BRIEF}~\citep{DBLP:conf/naacl/LiGWCP25} is a T5-based context compressor that can only handle a maximum sequence length of 512 tokens at a time. To accommodate this limitation, long contexts were uniformly divided into chunks of up to 512 tokens. Each chunk was then compressed using the trained compressor, and the compressed results of each chunk were concatenated to form the overall compressed summary.
    \textbullet{} \underline{Llama-3.1-3B-Instruct} denotes the off-the-shelf official release without further fine-tuning.
    \textbullet{} \underline{LongLLMLingua}~\citep{DBLP:conf/acl/JiangWL0L0Q24} uses LLMLingua~\citep{DBLP:conf/emnlp/JiangWLYQ23} as the backbone and further improves its perception of key information pertinent to the question. We followed their 2,000 tokens compression constraint. It performs both coarse-grained, demonstration-level compression and fine-grained, token-level compression, leveraging the perplexity of each demonstration or token calculated by a causal LM \texttt{Llama-2-7B-Chat}. 
    \textbullet{} \underline{GPT-4.1-nano} was prompted to summarize the documents with respect to the question.

  \subsection{Implementation} \label{sec-implement}
  The LoRA technique~\citep{DBLP:conf/iclr/HuSWALWWC22} was adopted for efficient compressor training on our curated data for three epochs.
  AdamW~\citep{DBLP:conf/iclr/LoshchilovH19} was used as the optimizer, and the batch size was set to 64.
  We utilize the Axolotl library~\citep{Axolotl} for efficient parallel training. 
  The entire training process takes about two days on 2 × NVIDIA A100 80GB GPUs.
  The Llama models were accessed via HuggingFace\footnote{\url{https://huggingface.co/meta-llama/Llama-3.1-8B-Instruct}} \footnote{\url{https://huggingface.co/meta-llama/Llama-3.1-70B-Instruct}} and GPT-4.1 was accessed via OpenAI\footnote{\url{https://platform.openai.com}}.

  \clearpage
  
  \subsection{More Experimental Results and Analysis} \label{sec-results}

\input{table/tab_diverse_domain}

  Table~\ref{tab:main-results-OOD} presents the evaluation results on three non-Wikipedia-based QA datasets across diverse domains using three different reader models. The selected datasets are the extended NarrativeQA (literature and film)~\citep{narrativeqa}, Qasper (science)~\citep{qasper}, and MultiFieldQA-en (multiple fields)~\citep{DBLP:conf/acl/BaiLZL0HDLZHDTL24} from LongBench~\citep{DBLP:conf/acl/BaiLZL0HDLZHDTL24}, with context lengths of 18.4k, 3.6k, and 4.5k words, respectively.
  BRIEF was not included because it was not designed for such diverse evaluations.

  \MODELNAME{} still demonstrates promising performance in this evaluation. Compared to other compression methods, \MODELNAME{}-\textsc{Auto} achieves the best QA performance. Notably, \MODELNAME{}-\textsc{High} attains the highest compression rate while still outperforming other compression baselines.
  Nevertheless, all compression methods fall short of the non-compression baseline in QA performance on these datasets, highlighting the need for further advances in robust compression across diverse domains.

  \vspace{1cm}
  \input{table/tab_expand_oracle_only_full}

  \clearpage

    \begin{figure*}[!htb]
      \centering
      \includegraphics[width=0.88\textwidth]{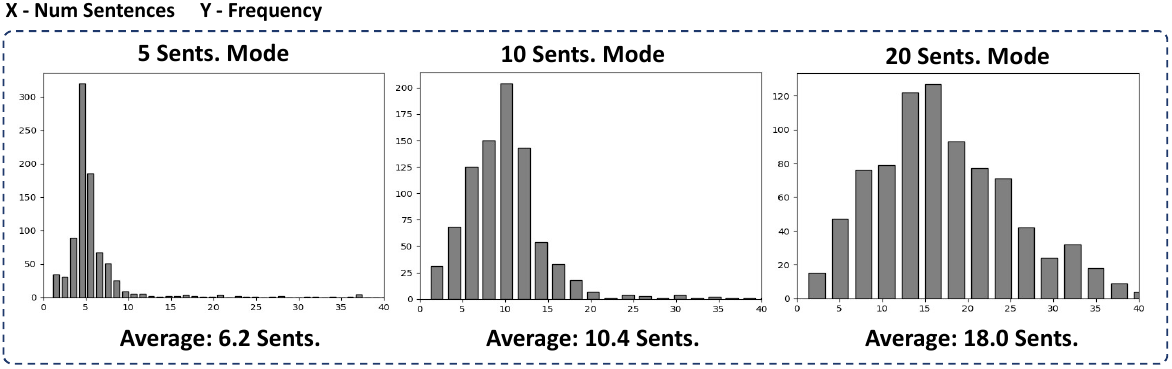}
      \caption{The distribution of sentence counts in the generated summary over four test sets in various compression modes. }
      \label{fig-user-controllable}
      \vspace{-2mm}
    \end{figure*}

\begin{figure*}[!htb]
  \centering
  \includegraphics[width=0.88\textwidth]{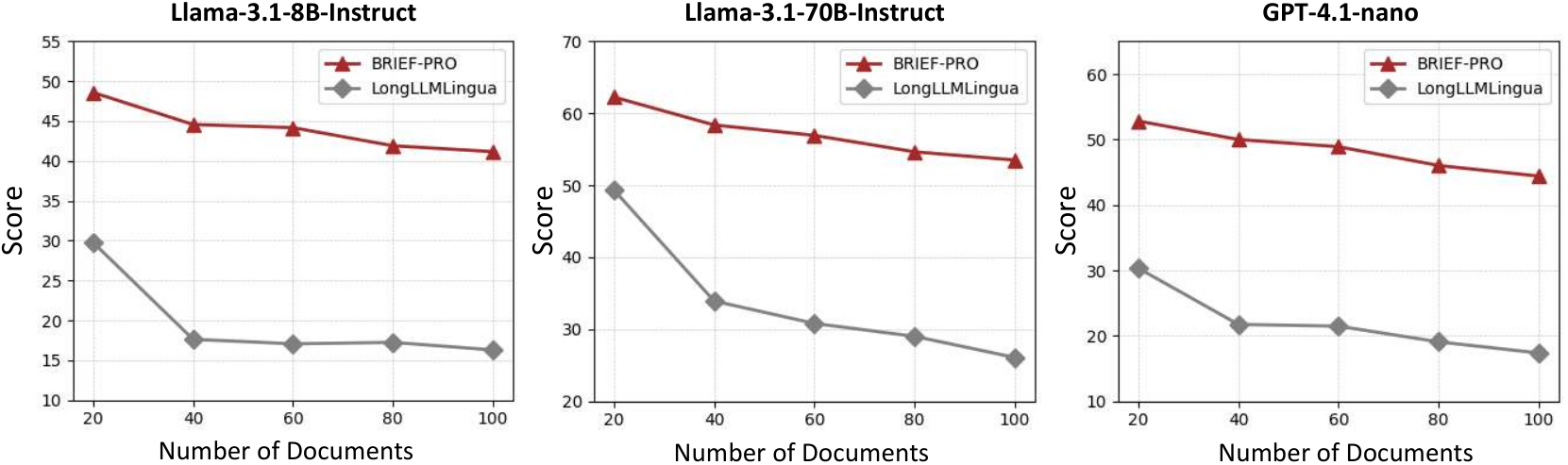}
  \caption{The performance of compressors under different context length. We expand the scope of retrieved documents from the top 20 to the top 100 based on the validation set in the Musique dataset~\citep{DBLP:journals/tacl/TrivediBKS22}. For BRIEF-PRO, we follow the AUTO setting. For LongLLMLingua~\citep{DBLP:conf/acl/JiangWL0L0Q24}, we follow their 2,000-token compression constraint. The reported score is the average of the F1 and EM scores.}
  \label{fig-context-length}
  \vspace{-2mm}
\end{figure*}

\begin{figure*}[!htb]
  \centering
  \includegraphics[width=0.98\textwidth]{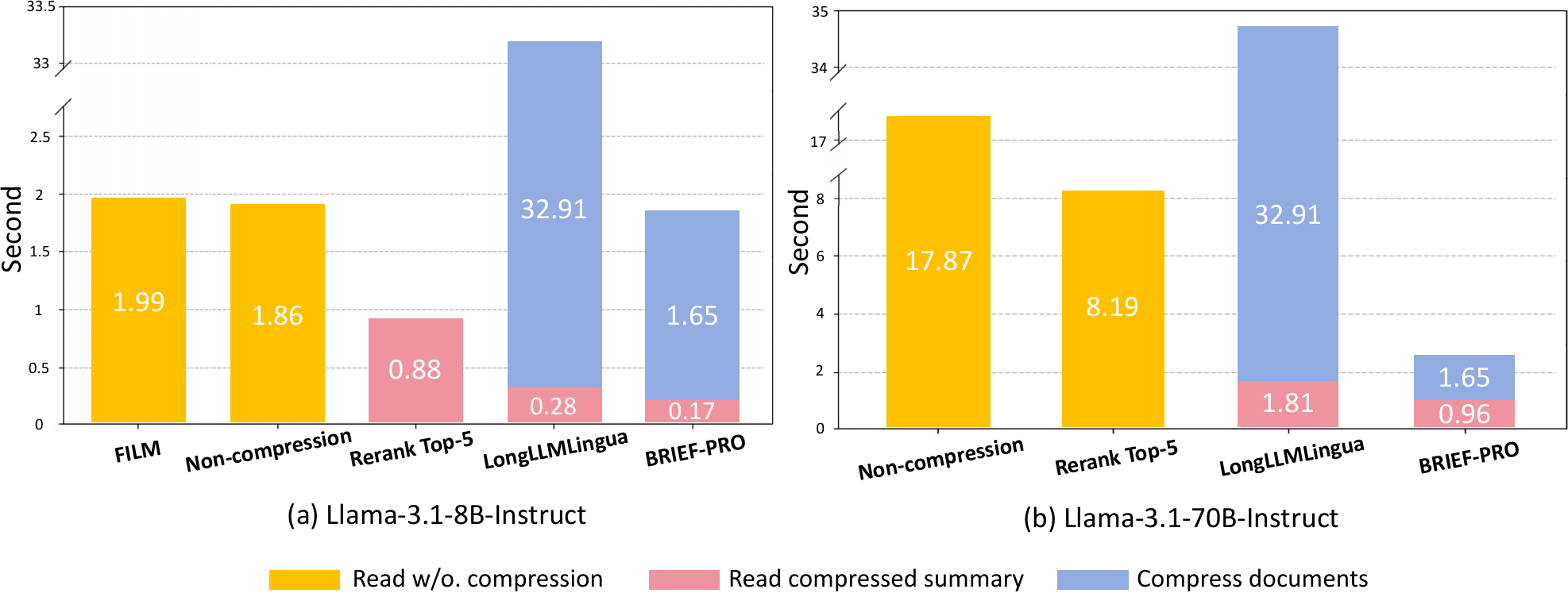}
  \caption{The comparison of end-to-end latency using (a) \texttt{Llama-3.1-8B-Instruct} and (b) \texttt{Llama-3.1-70B-Instruct} as the reader model. For each method, we computed the total running time by summing the execution times of all components in the pipeline. 
  Experimental results demonstrate that \MODELNAME{} consistently reduces the overall end-to-end latency. 
  Notably, on the 70B model, latency is reduced to only 14\%, 32\%, and 7\% of that of No-Compression, Rerank Top-5, and LongLLMLingua, respectively.}
  \label{latency_8B}
  \vspace{-2mm}
\end{figure*}

\clearpage

\subsection{Qualitative Examples} \label{sec-qualitative_examples}

We provide qualitative examples of correct and incorrect compression cases. Sentences highlighted in blue indicate key information for answering the question.

In the correct case, this is a two-hop question: the key information is distributed across two passages in an extremely long context, requiring the compression model to identify sub-questions and precisely locate the relevant sentences. \MODELNAME{} performs well at locating key information in such long contexts, thanks to the high-quality training data synthesized by our method, which is highly representative of this setting.

In the incorrect case, \MODELNAME{} shows a typical mistake in context compression for multi-hop questions: it fails to capture relevant information in one of the hops. This problem is common across existing methods and becomes more severe with long contexts, where locating the relevant information is much harder. It reflects a broader challenge for the field and warrants further investigation.

\begin{tcolorbox}[
    colframe=black, 
    colback=white, 
    coltitle=white, 
    colbacktitle=black, 
    title=Correct Case (10343 words -> 282 words | Compression rate: 36x),
    fonttitle=\bfseries,
    boxrule=1pt,
    arc=2mm,    
    left=4mm, right=4mm, top=3mm, bottom=3mm, 
    enhanced, 
    breakable
]
{
\setlength{\parskip}{5pt}

\textbullet\ \textbf{Question:}
Robbie Tucker plays in what series that follows a group of friends who run an Irish bar?

\vspace{8pt}

\textbullet\ \textbf{Context Before Compression (10343 words):}

Passage 1:
John Franks (judge)
Sir John Franks (1770–1852), was an Indian judge.
Franks was the second son of Thomas Franks (1729–1787), of Ballymagooly, County Cork, by Catherine, daughter of Rev. John Day. He was born in 1770, and graduated at Trinity College, Dublin, B.A. 1788, LL.B. 1791. He was called to the Irish Bar 1792. He went on the Munster circuit, and had a good practice as chamber counsel.

...

\textbf{\textcolor{mydarkblue}{Robbie Tucker (born April 5, 2001) is an American actor.}} His best known role to date is that of Fenmore Baldwin on the CBS soap opera The Young and the Restless. \textbf{\textcolor{mydarkblue}{Tucker has also starred on other series, such as Criminal Minds, FlashForward and It's Always Sunny in Philadelphia.}} He has also appeared in the films Prom and Little Fockers.
In 2012, Tucker was nominated at the 33rd Young Artist Awards for his performance in Prom and won for his role in The Young and the Restless.He is also the brother of actress Jillian Rose Reed.

Filmography

...

\textbf{\textcolor{mydarkblue}{It's Always Sunny in Philadelphia is an American sitcom}} created by Rob McElhenney and developed with Glenn Howerton for FX. It premiered on August 4, 2005, and was moved to FXX beginning with the ninth season in 2013. It stars Charlie Day, Howerton, McElhenney, Kaitlin Olson, and Danny DeVito. The series follows the exploits of "The Gang", \textbf{\textcolor{mydarkblue}{a group of narcissistic and sociopathic friends who run the Irish dive bar Paddy's Pub in South Philadelphia}}, Pennsylvania, but spend most of their free time drinking, scheming, arguing amongst themselves, and plotting elaborate cons against others, and at times each other, for personal benefit, financial gain, revenge, or simply out of boredom.
With its 15th season, the series became the longest-running live-action American comedy series, replacing The Adventures of Ozzie and Harriet, which ran for 14 seasons (435 episodes) between 1952 and 1966.

...

Podcast
On November 9, 2021, Howerton, Day, and McElhenney started The Always Sunny Podcast, an episode-by-episode-recap of the show, with Megan Ganz as producer. 
Occasional guest stars include Kaitlin Olson, David Hornsby, Mary Elizabeth Ellis, Michael Naughton, Lin-Manuel Miranda, Cormac Bluestone and Danny DeVito.

\vspace{8pt}

\textbullet\ \textbf{Context After Compression (282 words):}

\textbf{\textcolor{mydarkblue}{Robbie Tucker (born April 5, 2001) is an American actor.}} His best known role to date is that of Fenmore Baldwin on the CBS soap opera The Young and the Restless. \textbf{\textcolor{mydarkblue}{Tucker has also starred on other series, such as Criminal Minds, FlashForward and It's Always Sunny in Philadelphia.}} He has also appeared in the films Prom and Little Fockers.
In 2012, Tucker was nominated at the 33rd Young Artist Awards for his performance in Prom and won for his role in The Young and the Restless.He is also the brother of actress Jillian Rose Reed.

\textbf{\textcolor{mydarkblue}{It's Always Sunny in Philadelphia is an American sitcom}} created by Rob McElhenney and developed with Glenn Howerton for FX. It premiered on August 4, 2005, and was moved to FXX beginning with the ninth season in 2013. It stars Charlie Day, Howerton, McElhenney, Kaitlin Olson, and Danny DeVito. The series follows the exploits of "The Gang", \textbf{\textcolor{mydarkblue}{a group of narcissistic and sociopathic friends who run the Irish dive bar Paddy's Pub in South Philadelphia,}} Pennsylvania, but spend most of their free time drinking, scheming, arguing amongst themselves, and plotting elaborate cons against others, and at times each other, for personal benefit, financial gain, revenge, or simply out of boredom.
With its 15th season, the series became the longest-running live-action American comedy series, replacing The Adventures of Ozzie and Harriet, which ran for 14 seasons (435 episodes) between 1952 and 1966. In December 2020, the series was renewed for a total of four additional seasons, bringing it to 18 seasons. The 16th season premiered on June 7, 2023.The show has received critical acclaim, with many lauding the cast performances and dark humor. It has amassed a large cult following.

\vspace{8pt}

\textbullet\ \textbf{Answer:}
It's Always Sunny in Philadelphia (\ding{51})

\textbullet\ \textbf{Ground Truth Answer:}
It's Always Sunny in Philadelphia

}
\end{tcolorbox}

\begin{tcolorbox}[
    colframe=black, 
    colback=white, 
    coltitle=white, 
    colbacktitle=black, 
    title=Incorrect Case (11179 words -> 140 words | Compression rate: 80x),
    fonttitle=\bfseries,
    boxrule=1pt,
    arc=2mm,    
    left=4mm, right=4mm, top=3mm, bottom=3mm, 
    enhanced, 
    breakable
]
{
\setlength{\parskip}{5pt}

\textbullet\ \textbf{Question:}
Roger Stuart Woolhouse is a biographer of a philosopher commonly known as what?

\vspace{8pt}

\textbullet\ \textbf{Context Before Compression (11179 words):}

Passage 1:
Philip Carlo
Philip Carlo (April 18, 1949 – November 8, 2010) was an American journalist and best selling biographer of Thomas Pitera, Richard Kuklinski, Anthony Casso, and Richard Ramirez. Carlo had amyotrophic lateral sclerosis (ALS), commonly known as "Lou Gehrig's Disease".

...

\textbf{\textcolor{mydarkblue}{John Locke  (; 29 August 1632 – 28 October 1704) was an English philosopher and physician,}} widely regarded as one of the most influential of Enlightenment thinkers and \textbf{\textcolor{mydarkblue}{commonly known as the "father of liberalism".}} Considered one of the first of the British empiricists, following the tradition of Francis Bacon, Locke is equally important to social contract theory. His work greatly affected the development of epistemology and political philosophy. His writings influenced Voltaire and Jean-Jacques Rousseau, and many Scottish Enlightenment thinkers, as well as the American Revolutionaries. His contributions to classical republicanism and liberal theory are reflected in the United States Declaration of Independence. Internationally, Locke's political-legal principles continue to have a profound influence on the theory and practice of limited representative government and the protection of basic rights and freedoms under the rule of law.Locke's theory of mind is often cited as the origin of modern conceptions of identity and the self, figuring prominently in the work of later philosophers such as Jean-Jacques Rousseau, David Hume, and Immanuel Kant.

...

\textbf{\textcolor{mydarkblue}{Roger Stuart Woolhouse (1940–2011) was an English philosopher, an expert on empiricism and rationalism and a biographer of John Locke.}}
He was born in Wath-upon-Dearne and educated at Saltburn Primary School, Sir William Turner's Grammar School, London University (Philosophy) and then Selwyn College, Cambridge for his Doctorate.From 1969 until his retirement in 2001, Woolhouse worked in the Department of Philosophy at the University of York.Cambridge University Press requested Woolhouse write a biography of Locke, the last major biography being Maurice Cranston's 1957 work. Woolhouse's biography appeared in 2007.
After his death, York's Department of Philosophy founded the Roger Woolhouse Prize, an annual £500 prize awarded to MA Philosophy students.

...

Descartes, Spinoza, Leibniz: The Concept of Substance in Seventeenth Century Metaphysics (Routledge, 1993).
Locke: A Biography (Cambridge University Press, 2007).
Starting with Leibniz (Continuum, 2010).

\vspace{8pt}

\textbullet\ \textbf{Context After Compression (140 words):}

\textbf{\textcolor{mydarkblue}{Roger Stuart Woolhouse (1940–2011) was an English philosopher, an expert on empiricism and rationalism and a biographer of John Locke.}}
He was born in Wath-upon-Dearne and educated at Saltburn Primary School, Sir William Turner's Grammar School, London University (Philosophy) and then Selwyn College, Cambridge for his Doctorate.From 1969 until his retirement in 2001, Woolhouse worked in the Department of Philosophy at the University of York.Cambridge University Press requested Woolhouse write a biography of Locke, the last major biography being Maurice Cranston's 1957 work. Woolhouse's biography appeared in 2007.
After his death, York's Department of Philosophy founded the Roger Woolhouse Prize, an annual £500 prize awarded to MA Philosophy students.

Works
Locke (Prentice Hall / Harvester Wheatsheaf, 1984).
The Empiricists (Oxford University Press, 1988).
Descartes, Spinoza, Leibniz: The Concept of Substance in Seventeenth Century Metaphysics (Routledge, 1993).
Locke: A Biography (Cambridge University Press, 2007).
Starting with Leibniz (Continuum, 2010).

\vspace{8pt}

\textbullet\ \textbf{Answer:}
John Locke (\ding{56})

\textbullet\ \textbf{Ground Truth Answer:}
Father of Liberalism

}
\end{tcolorbox}

\section{Discussions}

  \subsection{Hallucination Risk}
  We would like to raise the discussion about the risk of hallucinations in abstractive and extractive compression. While extractive methods avoid generating new tokens, they are not immune to hallucinations. In particular, by removing essential context, extractive compressors may yield misleading interpretations; by elevating marginal sentences, they may distort relevance; and by retaining unresolved references (e.g., pronouns, temporal markers), they may create spurious associations. Thus, both extractive and abstractive approaches carry inherent risks, though manifested in different ways. We acknowledge and share the ambition that mitigating hallucinations should be a key objective, regardless of whether compressors are abstractive or extractive.

  \subsection{Importance of Training a Long-context Compressor}
  The goal of long-context compression is to reduce inference cost without degrading downstream performance. While modern LLMs can perform aspect-based and length-controlled summarization, using them directly for compression increases inference cost and undermines this objective. Our focus extends beyond absolute context length to include efficiency, cost, and accessibility. Specifically, our contribution is to train a lightweight compression model that achieves strong long-context compression at low inference cost while preserving task quality, which is a non-trivial objective that requires careful training design rather than simply invoking modern LLMs. Long-context compression can be positioned as a complementary approach to future long-context models, enhancing efficiency and reasoning robustness rather than merely extending the maximum token length.
  
  Furthermore, our experimental results show that fine-tuning a compression model for these tasks is also non-trivial. The core challenge lies in constructing effective training data. Our preliminary experiments show that the definition of the target summary strongly influences both training difficulty and model performance. For example, we currently adopt the Head-Tail Iterative Pruning method because removing unhelpful sentences in the middle produces multiple discrete text segments rather than a coherent and continuous summary. While this approach can yield more concise outputs, it substantially increases training difficulty and results in suboptimal model performance. These findings reinforce our belief that the training method proposed in this work makes an important contribution to efficient compression model training.

  \subsection{Explanation of Head-Tail Assumption}
  We agree that critical information can be distributed across multiple sentences, and this does not contradict our implementation. For example, if important content appears at the beginning or end of a document, our proposed Head-Tail Iterative Pruning method will halt pruning and retain those segments. Our design aims to identify a more compact, continuous, and informative text span within each oracle document. By retaining a continuous central segment, our approach captures the majority of salient information while maintaining textual coherence. This design is motivated by our preliminary experiments showing that removing middle sentences often produces fragmented and disconnected spans, which significantly increases the difficulty of compression training. Pruning head and tail sentences thus offers a practical balance, preserving continuity while focusing on the most informative portion of the document.

  \subsection{Head-to-Head Comparison against GPT-4.1-nano}
  While GPT-4.1-nano shows strong performance across benchmarks, our focus with BRIEF-Pro is on democratizing LLM training and inference. BRIEF-Pro enables high-quality compression using smaller, open-weight models, making it accessible to researchers and practitioners who do not have access to proprietary or extremely large LLMs. Training a compressor, as required by BRIEF-Pro, is a one-time and extremely low cost that allows repeated efficient inference on downstream tasks without relying on expensive API calls or large-scale models. In this sense, BRIEF-Pro prioritizes accessibility, reproducibility, and computational efficiency, complementing rather than directly competing with very large, closed-weight LLMs like GPT-4.1.

  \subsection{Scale across Multiple Retrievals}
  BRIEF-Pro primarily focuses on single-retrieval compression, with key contributions including user-controlled compression and support for longer contexts. While our current experiments are limited to single-retrieval settings, BRIEF-Pro is modular and scalable, allowing each retrieval to be compressed individually before aggregation in multi-retrieval scenarios. We believe this approach can be seamlessly integrated into iterative retrieval pipelines, and exploring such multi-retrieval applications is an important direction for future work.

  \subsection{Clarification on the Definition of Sentence Usefulness and Model Dependence}
  In our definition of helpfulness, we used Llama-3-8B-Instruct to compute log-likelihoods. This model was not cherry-picked to maximize our final results. Instead, we simply adopted a commonly used open-source model without tuning this choice.
  
  For each sentence, we recompute the log-likelihood of the correct answer after removing that sentence, and we label the sentence as unhelpful and remove it if and only if this removal makes the correct answer more likely. In other words, the threshold is simply “remove if the log-likelihood increases,” and no extra hyperparameters are introduced or tuned.
  
  We acknowledge that the likelihood results and the final constructed data might be model-specific, but the model is used only once offline to construct the training supervision. Our experiments show that using data constructed with this single, fixed model already yields consistent gains across multiple, diverse reader models, including black-box ones, suggesting that the induced notion of usefulness generalizes beyond the specific LM used for data synthesis.

\subsection{Data Contamination Analysis}

\begin{table}[h!]

\centering
\renewcommand{\arraystretch}{1}
\begin{adjustbox}{width=\textwidth}
\begin{tabular}{
  >{\centering\arraybackslash}p{2.8cm} 
  >{\centering\arraybackslash}p{2.8cm}
  >{\centering\arraybackslash}p{2.8cm} 
  >{\centering\arraybackslash}p{2.8cm} 
  >{\centering\arraybackslash}p{2.8cm} 
  >{\centering\arraybackslash}p{2.8cm} 
  >{\centering\arraybackslash}p{2.8cm} 
}

\toprule

MuSiqQue & HotpotQA & 2WikiMultiHopQA & LongSeal & NarrativeQA & Qasper & MultiFieldQA-en  \\
\midrule

0.095 & 0.086 & 0.029 & 0.008 & 0.0007 & 0.002 & 0.062 \\

\bottomrule
\end{tabular}
\end{adjustbox}
\caption{Average per-sample maximum 5-gram overlap between each test set and the training corpus.}
\label{Table:contamination-study}
\end{table}

Since our training pipeline uses Wikipedia for data expansion, one potential concern is the possibility of overlap between the expanded training data and the test sets, which could lead to data leakage.
We would like to clarify that no data leakage occurs during compressor training. The compressor is query-aware, and each training example consists of a document context, a query, and an oracle text (i.e., the compressor output). The queries and oracle texts in the training set do not overlap with those in the test set, preventing the compressor from observing identical query–oracle compression patterns during training.

To provide a more rigorous contamination analysis, we further quantify the potential overlap between document contexts in the training and test sets. For each test example, we compute its 5-gram overlap with every training example, take the maximum overlap score, and then report the average of these maxima for each dataset. The 5-gram overlap is computed as follows:

\begin{equation}
\mathrm{Overlap}(\mathrm{context}_{test}, \mathrm{context}_{train}) \;=\; \frac{|\mathcal{G}_5(\mathrm{context}_{test}) \cap \mathcal{G}_5(\mathrm{context}_{train})|}{|\mathcal{G}_5(\mathrm{context}_{test})|},
\end{equation}
where $\mathcal{G}_5(x)$ denotes the set of unique 5-grams (contiguous 5-word shingles) extracted from text $x$ after normalization and tokenization.

Table~\ref{Table:contamination-study} presents the 5-gram overlap results. Across all datasets, the average maximum 5-gram overlap is below 0.1, and for some datasets it is below 0.01, indicating only slight (or nearly no) contextual overlap between the training and test sets. Therefore, the training-test contamination in our expanded training data is negligible and unlikely to affect our results.

%% file: table/tab_diverse_domain.tex
\begin{table*}[ht!]
\centering
\resizebox{\linewidth}{!}{
\def\arraystretch{1.1}
\begin{tabular}{llcccccccccccccc}
\toprule
  \multirow{2}{*}{\textbf{Category}} & \multirow{2}{*}{\textbf{Method}} &  \multicolumn{3}{c}{\textbf{NarrativeQA}}   & \multicolumn{3}{c}{\textbf{Qasper}} & \multicolumn{3}{c}{\textbf{MultiFieldQA-en}} & \multicolumn{2}{c}{\textbf{Average}} \\
                                   & & \textbf{EM} & \textbf{F1} & \textbf{Rate} & \textbf{EM} & \textbf{F1} & \textbf{Rate} & \textbf{EM} & \textbf{F1} & \textbf{Rate} & \textbf{QA} & \textbf{Rate} \\
\midrule
 \multirow{2}{*}{Long-context LLMs} & ProLong-8B & 14 & 31.63 & 1x & 18.5 & 28.78 & 1x & 22 & 51.44 & 1x & 27.73 & 1x \\
                                    & FILM-7B & 10 & 28.47 & 1x & 21 & 45.25 & 1x & 22 & 54.34 & 1x & 30.18 &  1x\\
\midrule
\rowcolor{gray!20} \multicolumn{16}{c}{\textbf{\texttt{Llama-3.1-8B-Instruct}}} \\
\midrule
  \multirow{1}{*}{Non-compression} & & 7.50 & 29.43 & 1x & 20.50 & 47.62 & 1x & 21.33 & 52.77 & 1x & 29.86 & 1x  \\
\cdashline{1-16}
 \multirow{3}{*}{Extractive}  

                          & RECOMP (Extractive) & 4.50 & 16.19 & 94x & 13.50 & 33.14 & 16x & \underline{17.33} & 43.82 & 16x & 21.41 & 42x \\

                          & EXIT &  \underline{5.50} & 20.11 & 40x & 13.00 & 24.28 & 15x & 10.67 & 25.33 & 14x & 16.48 & 23x  \\

                          & Rerank Top-3 & 5.00 & 20.06 & 3x & 14.00 & 39.31 & 3x & 16.50 & 43.10 & 3x & 23.00 &  3x  \\
                          
\cdashline{1-16}
 \multirow{9}{*}{Abstractive} & RECOMP (Abstractive) & 3.50 & 13.69 & 17x & 12.00 & 20.52 & 17x & 8.00 & 30.14 & 12x & 14.64 & 15x\\

                              & Llama-3.2-3B-Instruct & \underline{5.50} & 19.81 & 45x & 11.50 & 29.28 & 12x & 14.67 & 42.30 & 16x & 20.51 & 24x \\
 
                              & LongLLMLingua & 4.50 & 21.56 & 17x &  \underline{14.50} &  \underline{41.18} & 3x &  \underline{17.33} &  \underline{44.62} & 4x & \underline{23.95} & 8x \\
                              
                              & GPT-4.1-nano & \textbf{6.50} & \textbf{24.46} & 91x & 13.50 & 35.89 & 19x & 14.67 & 41.95 & 29x & 22.83 & 46x \\
                              
                              & \MODELNAME{}-\textsc{Auto} &  \underline{5.50} &  \underline{21.73} & 32x & \textbf{16.50} & \textbf{42.70} & 7x & \textbf{24.67} & \textbf{54.21} & 11x & \textbf{27.55} &  17x \\

                              \cdashline{2-16}

                              & \MODELNAME{}-\textsc{High} & 6.50 & 19.10 & 97x & 14.50 & 36.77 & 22x & 24.00 & 52.43 & 29x & 25.55 &  49x\\

                              & \MODELNAME{}-\textsc{Medium} & 6.50 & 20.40 & 66x & 14.50 & 40.95 & 12x & 24.00 & 52.23 & 16x & 26.43 & 31x\\
 
                              & \MODELNAME{}-\textsc{Low} & 7.00 & 21.84 & 40x & 15.00 & 40.50 & 7x & 24.00 & 53.88 & 10x & 27.04 &  19x\\

 \midrule
\rowcolor{gray!20} \multicolumn{16}{c}{\textbf{\texttt{Llama-3.1-70B-Instruct}}} \\
\midrule
  \multirow{1}{*}{Non-compression} & & 15.00 & 35.20 &  1x & 21.00 & 48.28 & 1x & 21.33 & 54.38 & 1x & 32.53  & 1x\\
\cdashline{1-16}
 \multirow{3}{*}{Extractive}  

                          & RECOMP (Extractive) & 6.00 & 16.62 & 94x & \underline{18.50} & 38.53 & 16x & \underline{18.00} & \underline{46.43} & 16x & 24.01 & 42x \\

                          & EXIT & \textbf{8.00} & 19.70 & 40x & 17.50 & 27.96 & 15x & 10.67 & 25.75 & 14x & 18.26 & 23x  \\

                          & Rerank Top-3 & \underline{7.50} & 21.62 & 3x & \underline{18.50} & 37.58 & 3x & 17.33 & 45.10 & 3x & 24.61 & 3x  \\

\cdashline{1-16}
 \multirow{9}{*}{Abstractive} & RECOMP (Abstractive) & 5.50 & 17.16 & 17x & 12.00 & 22.86 & 17x & 11.33 & 33.78 & 12x & 17.11 & 15x \\

                              & Llama-3.2-3B-Instruct & 7.00 & 19.95 & 45x & 16.50 & 35.23 & 12x & 12.80 & 39.31 & 16x & 21.80 & 24x \\

                              & LongLLMLingua & \textbf{8.00} & 21.09 & 17x & 18.00 & \underline{45.26} & 3x & 16.67 & 45.87 & 4x & \underline{25.82} & 8x \\

                              & GPT-4.1-nano & \underline{7.50} & \textbf{22.91} & 91x & 14.00 & 37.41 & 19x & 13.33 & 41.44 & 29x & 22.77 & 46x \\
                              
                              & \MODELNAME{}-\textsc{Auto} & \textbf{8.00} & \underline{22.83} & 32x & \textbf{22.00} & \textbf{47.40} & 7x & \textbf{20.67} & \textbf{51.16} & 11x & \textbf{28.68} &  17x\\

                              \cdashline{2-16}
                              
                              & \MODELNAME{}-\textsc{High} & 7.00 & 18.65 &  97x & 20.50 & 42.82 &  22x & 21.33 & 50.70 & 29x & 26.83 & 49x\\

                              & \MODELNAME{}-\textsc{Medium} & 6.50 & 20.02 & 66x & 19.50 & 46.16 & 12x & 20.67 & 51.93 & 16x & 27.46 & 31x\\
 
                              & \MODELNAME{}-\textsc{Low} & 7.50 & 20.71 & 40x & 19.00 & 46.37 & 7x & 20.67 & 50.63 & 10x & 27.48 &  19x\\
                              
\midrule
\rowcolor{gray!20} \multicolumn{16}{c}{\textbf{\texttt{GPT-4.1-nano}}} \\
\midrule
  \multirow{1}{*}{Non-compression} & & 10.50 & 28.25 & 1x & 20.50 & 44.47 & 1x & 16.67 & 50.09 & 1x & 28.41 & 1x \\
\cdashline{1-16}
 \multirow{3}{*}{Extractive}  

                          & RECOMP (Extractive) & 5.00 & 15.62 & 94x & \underline{17.50} & 37.46 & 16x & 12.67 & 43.72 & 16x & 22.00 & 42x \\

                          & EXIT & \underline{7.50} & 18.45 & 40x & 11.50 & 25.40 & 15x & 9.33 & 27.12 & 14x & 16.55 &  23x  \\

                          & Rerank Top-3 & \underline{7.50} & 19.10 & 3x & 16.00 & 38.04 & 3x & 13.33 & 42.67 & 3x & 22.77 & 3x  \\

\cdashline{1-16}
 \multirow{9}{*}{Abstractive} & RECOMP (Abstractive) & 5.50 & 14.01 & 17x & 8.00 & 18.32 & 17x & 7.33 & 28.83 & 12x & 13.67 & 15x \\

                              & Llama-3.2-3B-Instruct & 5.50 & 17.15 & 45x & 12.00 & 32.39 & 12x & 12.67 & 41.55 & 16x & 20.21 & 24x\\
 
                              & LongLLMLingua & 4.00 & 18.24 & 17x & 15.50 & \underline{39.53} & 3x & \underline{14.00} & \underline{44.88} & 4x & 22.69 & 8x \\
                              
                              & GPT-4.1-nano & \underline{7.50} & \textbf{24.50} & 91x & 16.50 & 38.64 & 19x & 12.67 & 41.04 & 29x & \underline{23.48} & 46x\\
                              
                              & \MODELNAME{}-\textsc{Auto} & \textbf{8.50} & \underline{19.53} & 32x & \textbf{19.00} & \textbf{42.21} & 7x & \textbf{16.67} & \textbf{48.17} & 11x & \textbf{25.68} &  17x \\

                              \cdashline{2-16}

                              & \MODELNAME{}-\textsc{High} & 7.50 & 18.88 & 97x & 16.50 & 38.78 & 22x & 18.00 & 48.53 & 29x & 24.70 &  49x \\

                              & \MODELNAME{}-\textsc{Medium} & 7.50 & 19.25 & 66x & 18.00 & 42.54 & 12x & 16.67 & 49.79 & 16x & 25.63 &  31x \\
 
                              & \MODELNAME{}-\textsc{Low} & 8.50 & 20.85 & 40x & 17.00 & 41.87 & 7x & 17.33 & 49.75 & 10x & 25.88 &  19x \\
                   
\bottomrule
\end{tabular}
}
\caption{Evaluation results on three non-Wikipedia-based QA datasets across diverse domains, using small, large, and proprietary LMs as the $\mathcal{M}$, respectively. 
\textbf{Bold} and \underline{underscore} denote the best and second-best QA performance, respectively, among different compression methods under the model’s self-determined compression setting (i.e., the \textsc{Auto} mode).
}

\label{tab:main-results-OOD}
\end{table*}

%% file: table/tab_expand_oracle_only_full.tex
\begin{table*}[!htb]
\centering
\resizebox{0.98\linewidth}{!}{
\def\arraystretch{1.1}
\begin{tabular}{lccccccccccccccc}
\toprule
  \multirow{2}{*}{\textbf{Training Data}} & \multirow{2}{*}{\textbf{Avg. Length}} & \multicolumn{3}{c}{\textbf{MuSiqQue}}   & \multicolumn{3}{c}{\textbf{HotpotQA}} & \multicolumn{3}{c}{\textbf{2WikiMultiHopQA}} & \multicolumn{3}{c}{\textbf{LongSeal}} & \multicolumn{2}{c}{\textbf{Average}} \\
                                   & & \textbf{EM} & \textbf{F1} & \textbf{Rate} & \textbf{EM} & \textbf{F1} & \textbf{Rate} & \textbf{EM} & \textbf{F1} & \textbf{Rate} & \textbf{EM} & \textbf{F1} & \textbf{Rate} & \textbf{QA} & \textbf{Rate} \\
\midrule

\rowcolor{gray!20} \multicolumn{16}{c}{\textbf{\texttt{Llama-3.1-8B-Instruct}}} \\
\midrule
  Oracle+ \& Distractor+ & 6.0k & 27.50 & 36.64 & 35x & 49.00 & 63.05 & 34x & 47.50 & 56.00 & 23x & 12.99 & 17.62 & 36x & 38.79 & 32x\\
  Oracle+ \& Distractor+ & 3.6k & 27.00 & 33.84 & 36x & 43.00 & 56.93 & 39x & 43.50 & 50.94 & 23x & 13.78 & 19.18 & 39x & 36.02 & 34x\\
  Oracle++               & 3.6k & 22.50 & 29.71 & 42x & 42.00 & 55.37 & 37x & 39.50 & 48.01 & 23x & 14.17 & 18.81 & 39x & 33.76 & 35x\\
 \midrule
 
\rowcolor{gray!20} \multicolumn{16}{c}{\textbf{\texttt{Llama-3.1-70B-Instruct}}} \\
\midrule              
  Oracle+ \& Distractor+ & 6.0k & 38.50 & 48.84 & 35x & 53.00 & 67.29 & 34x & 59.50 & 66.94 & 23x & 12.99 & 17.59 & 36x & 45.58 & 32x\\
  Oracle+ \& Distractor+ & 3.6k & 36.50 & 45.07 & 36x & 47.00 & 60.97 & 39x & 51.50 & 60.23 & 23x & 13.78 & 18.90 & 39x & 41.74 & 34x\\
  Oracle++               & 3.6k & 32.50 & 41.64 & 42x & 48.50 & 61.67 & 37x & 52.50 & 62.33 & 23x & 14.57 & 19.73 & 39x & 41.68 & 35x\\                       
\midrule

\rowcolor{gray!20} \multicolumn{16}{c}{\textbf{\texttt{GPT-4.1-nano}}} \\
\midrule                   
  Oracle+ \& Distractor+ & 6.0k & 29.50 & 41.10 & 35x & 51.50 & 65.52 & 34x & 51.00 & 58.68 & 23x & 12.20 & 16.87 & 36x & 40.80 & 32x \\
  Oracle+ \& Distractor+ & 3.6k & 32.00 & 39.00 & 36x & 47.50 & 60.79 & 39x & 50.00 & 57.24 & 23x & 11.02 & 15.35 & 39x & 39.11 & 34x \\
  Oracle++               & 3.6k & 28.00 & 34.54 & 42x & 46.00 & 58.95 & 37x & 47.50 & 55.46 & 23x & 10.24 & 15.55 & 39x & 37.03 & 35x \\                     
\bottomrule
\end{tabular}
}
\caption{Evaluation results of comparing with exhaustively expanding only oracle documents.}

\label{tab-expand-oracle-only-full}
\vspace{-2mm}
\end{table*}

%% file: custom.bib
@article{DBLP:journals/csur/JiLFYSXIBMF23,
  author       = {Ziwei Ji and
                  Nayeon Lee and
                  Rita Frieske and
                  Tiezheng Yu and
                  Dan Su and
                  Yan Xu and
                  Etsuko Ishii and
                  Yejin Bang and
                  Andrea Madotto and
                  Pascale Fung},
  title        = {Survey of Hallucination in Natural Language Generation},
  journal      = {{ACM} Comput. Surv.},
  volume       = {55},
  number       = {12},
  pages        = {248:1--248:38},
  year         = {2023},
  url          = {https://doi.org/10.1145/3571730},
  doi          = {10.1145/3571730},
  bibsource    = {dblp computer science bibliography, https://dblp.org}
}

@inproceedings{DBLP:conf/nips/LewisPPPKGKLYR020,
  author       = {Patrick S. H. Lewis and
                  Ethan Perez and
                  Aleksandra Piktus and
                  Fabio Petroni and
                  Vladimir Karpukhin and
                  Naman Goyal and
                  Heinrich K{\"{u}}ttler and
                  Mike Lewis and
                  Wen{-}tau Yih and
                  Tim Rockt{\"{a}}schel and
                  Sebastian Riedel and
                  Douwe Kiela},
  editor       = {Hugo Larochelle and
                  Marc'Aurelio Ranzato and
                  Raia Hadsell and
                  Maria{-}Florina Balcan and
                  Hsuan{-}Tien Lin},
  title        = {Retrieval-Augmented Generation for Knowledge-Intensive {NLP} Tasks},
  booktitle    = {Advances in Neural Information Processing Systems 33: Annual Conference
                  on Neural Information Processing Systems 2020, NeurIPS 2020, December
                  6-12, 2020, virtual},
  year         = {2020},
  url          = {https://proceedings.neurips.cc/paper/2020/hash/6b493230205f780e1bc26945df7481e5-Abstract.html},
  bibsource    = {dblp computer science bibliography, https://dblp.org}
}

@inproceedings{DBLP:conf/emnlp/JiangWLYQ23,
  author       = {Huiqiang Jiang and
                  Qianhui Wu and
                  Chin{-}Yew Lin and
                  Yuqing Yang and
                  Lili Qiu},
  editor       = {Houda Bouamor and
                  Juan Pino and
                  Kalika Bali},
  title        = {LLMLingua: Compressing Prompts for Accelerated Inference of Large
                  Language Models},
  booktitle    = {Proceedings of the 2023 Conference on Empirical Methods in Natural
                  Language Processing, {EMNLP} 2023, Singapore, December 6-10, 2023},
  pages        = {13358--13376},
  publisher    = {Association for Computational Linguistics},
  year         = {2023},
  url          = {https://doi.org/10.18653/v1/2023.emnlp-main.825},
  doi          = {10.18653/V1/2023.EMNLP-MAIN.825},
  bibsource    = {dblp computer science bibliography, https://dblp.org}
}

@inproceedings{DBLP:conf/iclr/XuSC24,
  author       = {Fangyuan Xu and
                  Weijia Shi and
                  Eunsol Choi},
  title        = {{RECOMP:} Improving Retrieval-Augmented LMs with Context Compression
                  and Selective Augmentation},
  booktitle    = {The Twelfth International Conference on Learning Representations,
                  {ICLR} 2024, Vienna, Austria, May 7-11, 2024},
  publisher    = {OpenReview.net},
  year         = {2024},
  url          = {https://openreview.net/forum?id=mlJLVigNHp},
  bibsource    = {dblp computer science bibliography, https://dblp.org}
}

@inproceedings{DBLP:conf/acl/MallenAZDKH23,
  author       = {Alex Mallen and
                  Akari Asai and
                  Victor Zhong and
                  Rajarshi Das and
                  Daniel Khashabi and
                  Hannaneh Hajishirzi},
  editor       = {Anna Rogers and
                  Jordan L. Boyd{-}Graber and
                  Naoaki Okazaki},
  title        = {When Not to Trust Language Models: Investigating Effectiveness of
                  Parametric and Non-Parametric Memories},
  booktitle    = {Proceedings of the 61st Annual Meeting of the Association for Computational
                  Linguistics (Volume 1: Long Papers), {ACL} 2023, Toronto, Canada,
                  July 9-14, 2023},
  pages        = {9802--9822},
  publisher    = {Association for Computational Linguistics},
  year         = {2023},
  url          = {https://doi.org/10.18653/v1/2023.acl-long.546},
  doi          = {10.18653/V1/2023.ACL-LONG.546},
  bibsource    = {dblp computer science bibliography, https://dblp.org}
}

@inproceedings{DBLP:conf/icml/ShiCMSDCSZ23,
  author       = {Freda Shi and
                  Xinyun Chen and
                  Kanishka Misra and
                  Nathan Scales and
                  David Dohan and
                  Ed H. Chi and
                  Nathanael Sch{\"{a}}rli and
                  Denny Zhou},
  editor       = {Andreas Krause and
                  Emma Brunskill and
                  Kyunghyun Cho and
                  Barbara Engelhardt and
                  Sivan Sabato and
                  Jonathan Scarlett},
  title        = {Large Language Models Can Be Easily Distracted by Irrelevant Context},
  booktitle    = {International Conference on Machine Learning, {ICML} 2023, 23-29 July
                  2023, Honolulu, Hawaii, {USA}},
  series       = {Proceedings of Machine Learning Research},
  volume       = {202},
  pages        = {31210--31227},
  publisher    = {{PMLR}},
  year         = {2023},
  url          = {https://proceedings.mlr.press/v202/shi23a.html},
  bibsource    = {dblp computer science bibliography, https://dblp.org}
}

@inproceedings{DBLP:conf/nips/Mu0G23,
  author       = {Jesse Mu and
                  Xiang Li and
                  Noah D. Goodman},
  editor       = {Alice Oh and
                  Tristan Naumann and
                  Amir Globerson and
                  Kate Saenko and
                  Moritz Hardt and
                  Sergey Levine},
  title        = {Learning to Compress Prompts with Gist Tokens},
  booktitle    = {Advances in Neural Information Processing Systems 36: Annual Conference
                  on Neural Information Processing Systems 2023, NeurIPS 2023, New Orleans,
                  LA, USA, December 10 - 16, 2023},
  year         = {2023},
  url          = {http://papers.nips.cc/paper\_files/paper/2023/hash/3d77c6dcc7f143aa2154e7f4d5e22d68-Abstract-Conference.html},
  bibsource    = {dblp computer science bibliography, https://dblp.org}
}

@inproceedings{DBLP:conf/emnlp/ChevalierWAC23,
  author       = {Alexis Chevalier and
                  Alexander Wettig and
                  Anirudh Ajith and
                  Danqi Chen},
  editor       = {Houda Bouamor and
                  Juan Pino and
                  Kalika Bali},
  title        = {Adapting Language Models to Compress Contexts},
  booktitle    = {Proceedings of the 2023 Conference on Empirical Methods in Natural
                  Language Processing, {EMNLP} 2023, Singapore, December 6-10, 2023},
  pages        = {3829--3846},
  publisher    = {Association for Computational Linguistics},
  year         = {2023},
  url          = {https://doi.org/10.18653/v1/2023.emnlp-main.232},
  doi          = {10.18653/V1/2023.EMNLP-MAIN.232},
  bibsource    = {dblp computer science bibliography, https://dblp.org}
}

@inproceedings{DBLP:conf/emnlp/Yang0ZBCSM18,
  author       = {Zhilin Yang and
                  Peng Qi and
                  Saizheng Zhang and
                  Yoshua Bengio and
                  William W. Cohen and
                  Ruslan Salakhutdinov and
                  Christopher D. Manning},
  editor       = {Ellen Riloff and
                  David Chiang and
                  Julia Hockenmaier and
                  Jun'ichi Tsujii},
  title        = {HotpotQA: {A} Dataset for Diverse, Explainable Multi-hop Question
                  Answering},
  booktitle    = {Proceedings of the 2018 Conference on Empirical Methods in Natural
                  Language Processing, Brussels, Belgium, October 31 - November 4, 2018},
  pages        = {2369--2380},
  publisher    = {Association for Computational Linguistics},
  year         = {2018},
  url          = {https://doi.org/10.18653/v1/d18-1259},
  doi          = {10.18653/V1/D18-1259},
  bibsource    = {dblp computer science bibliography, https://dblp.org}
}

@inproceedings{DBLP:conf/emnlp/KarpukhinOMLWEC20,
  author       = {Vladimir Karpukhin and
                  Barlas Oguz and
                  Sewon Min and
                  Patrick S. H. Lewis and
                  Ledell Wu and
                  Sergey Edunov and
                  Danqi Chen and
                  Wen{-}tau Yih},
  editor       = {Bonnie Webber and
                  Trevor Cohn and
                  Yulan He and
                  Yang Liu},
  title        = {Dense Passage Retrieval for Open-Domain Question Answering},
  booktitle    = {Proceedings of the 2020 Conference on Empirical Methods in Natural
                  Language Processing, {EMNLP} 2020, Online, November 16-20, 2020},
  pages        = {6769--6781},
  publisher    = {Association for Computational Linguistics},
  year         = {2020},
  url          = {https://doi.org/10.18653/v1/2020.emnlp-main.550},
  doi          = {10.18653/V1/2020.EMNLP-MAIN.550},
  bibsource    = {dblp computer science bibliography, https://dblp.org}
}

@article{DBLP:journals/tmlr/IzacardCHRBJG22,
  author       = {Gautier Izacard and
                  Mathilde Caron and
                  Lucas Hosseini and
                  Sebastian Riedel and
                  Piotr Bojanowski and
                  Armand Joulin and
                  Edouard Grave},
  title        = {Unsupervised Dense Information Retrieval with Contrastive Learning},
  journal      = {Trans. Mach. Learn. Res.},
  volume       = {2022},
  year         = {2022},
  url          = {https://openreview.net/forum?id=jKN1pXi7b0},
  bibsource    = {dblp computer science bibliography, https://dblp.org}
}

@inproceedings{DBLP:conf/iclr/0008PWM0LSBSC24,
  author       = {Peng Xu and
                  Wei Ping and
                  Xianchao Wu and
                  Lawrence McAfee and
                  Chen Zhu and
                  Zihan Liu and
                  Sandeep Subramanian and
                  Evelina Bakhturina and
                  Mohammad Shoeybi and
                  Bryan Catanzaro},
  title        = {Retrieval meets Long Context Large Language Models},
  booktitle    = {The Twelfth International Conference on Learning Representations,
                  {ICLR} 2024, Vienna, Austria, May 7-11, 2024},
  publisher    = {OpenReview.net},
  year         = {2024},
  url          = {https://openreview.net/forum?id=xw5nxFWMlo},
  bibsource    = {dblp computer science bibliography, https://dblp.org}
}

@article{DBLP:journals/tacl/LiuLHPBPL24,
  author       = {Nelson F. Liu and
                  Kevin Lin and
                  John Hewitt and
                  Ashwin Paranjape and
                  Michele Bevilacqua and
                  Fabio Petroni and
                  Percy Liang},
  title        = {Lost in the Middle: How Language Models Use Long Contexts},
  journal      = {Trans. Assoc. Comput. Linguistics},
  volume       = {12},
  pages        = {157--173},
  year         = {2024},
  url          = {https://doi.org/10.1162/tacl\_a\_00638},
  doi          = {10.1162/TACL\_A\_00638},
  bibsource    = {dblp computer science bibliography, https://dblp.org}
}

@article{DBLP:journals/tacl/TrivediBKS22,
  author       = {Harsh Trivedi and
                  Niranjan Balasubramanian and
                  Tushar Khot and
                  Ashish Sabharwal},
  title        = {{MuSiQue}: Multihop Questions via Single-hop Question
                  Composition},
  journal      = {Trans. Assoc. Comput. Linguistics},
  volume       = {10},
  pages        = {539--554},
  year         = {2022},
  url          = {https://doi.org/10.1162/tacl\_a\_00475},
  doi          = {10.1162/TACL\_A\_00475},
  bibsource    = {dblp computer science bibliography, https://dblp.org}
}

@inproceedings{DBLP:conf/coling/HoNSA20,
  author       = {Xanh Ho and
                  Anh{-}Khoa Duong Nguyen and
                  Saku Sugawara and
                  Akiko Aizawa},
  editor       = {Donia Scott and
                  N{\'{u}}ria Bel and
                  Chengqing Zong},
  title        = {Constructing {A} Multi-hop {QA} Dataset for Comprehensive Evaluation
                  of Reasoning Steps},
  booktitle    = {Proceedings of the 28th International Conference on Computational
                  Linguistics, {COLING} 2020, Barcelona, Spain (Online), December 8-13,
                  2020},
  pages        = {6609--6625},
  publisher    = {International Committee on Computational Linguistics},
  year         = {2020},
  url          = {https://doi.org/10.18653/v1/2020.coling-main.580},
  doi          = {10.18653/V1/2020.COLING-MAIN.580},
  bibsource    = {dblp computer science bibliography, https://dblp.org}
}

@inproceedings{DBLP:conf/emnlp/YoonLHJK24,
  author       = {Chanwoong Yoon and
                  Taewhoo Lee and
                  Hyeon Hwang and
                  Minbyul Jeong and
                  Jaewoo Kang},
  editor       = {Yaser Al{-}Onaizan and
                  Mohit Bansal and
                  Yun{-}Nung Chen},
  title        = {CompAct: Compressing Retrieved Documents Actively for Question Answering},
  booktitle    = {Proceedings of the 2024 Conference on Empirical Methods in Natural
                  Language Processing, {EMNLP} 2024, Miami, FL, USA, November 12-16,
                  2024},
  pages        = {21424--21439},
  publisher    = {Association for Computational Linguistics},
  year         = {2024},
  url          = {https://aclanthology.org/2024.emnlp-main.1194},
  bibsource    = {dblp computer science bibliography, https://dblp.org}
}

@misc{meta2024llama31,
  author       = {{Meta AI}},
  title        = {{Introducing Llama‑3.1: Our Most Capable Models to Date}},
  howpublished = {Blog post},
  year         = {2024},
  month        = jul,
  day          = {23},
  url          = {https://ai.meta.com/blog/meta-llama-3-1/}
}

@misc{meta2024llama3_2_release,
  author       = {{Meta AI}},
  title        = {Llama 3.2 Community License Release},
  howpublished = {Meta AI},
  month        = sep,
  year         = {2024},
  url          = {https://huggingface.co/meta-llama/Llama-3.2-1B}
}

@misc{anthropic2024claude35,
  author       = {{Anthropic}},
  title        = {{Introducing Claude 3.5 Sonnet}},
  year         = {2024},
  month        = jun,
  url          = {https://www.anthropic.com/news/claude-3-5-sonnet},
}

@misc{google2025gemini25,
  author       = {{Google}},
  title        = {{Start Building with Gemini 2.5 Flash}},
  year         = {2025},
  month        = apr,
  url          = {https://developers.googleblog.com/en/start-building-with-gemini-25-flash/}
}

@misc{openai2025gpt41,
  author       = {{OpenAI}},
  title        = {{Introducing GPT-4.1 in the API}},
  year         = {2025},
  month        = apr,
  url          = {https://openai.com/index/gpt-4-1/}
}

@inproceedings{DBLP:conf/naacl/LiGWCP25,
  author       = {Yuankai Li and
                  Jia{-}Chen Gu and
                  Di Wu and
                  Kai{-}Wei Chang and
                  Nanyun Peng},
  editor       = {Luis Chiruzzo and
                  Alan Ritter and
                  Lu Wang},
  title        = {{BRIEF:} Bridging Retrieval and Inference for Multi-hop Reasoning
                  via Compression},
  booktitle    = {Findings of the Association for Computational Linguistics: {NAACL}
                  2025, Albuquerque, New Mexico, USA, April 29 - May 4, 2025},
  pages        = {5449--5470},
  publisher    = {Association for Computational Linguistics},
  year         = {2025},
  url          = {https://doi.org/10.18653/v1/2025.findings-naacl.301},
  doi          = {10.18653/V1/2025.FINDINGS-NAACL.301},
  bibsource    = {dblp computer science bibliography, https://dblp.org}
}

@inproceedings{DBLP:conf/acl/BaiLZL0HDLZHDTL24,
  author       = {Yushi Bai and
                  Xin Lv and
                  Jiajie Zhang and
                  Hongchang Lyu and
                  Jiankai Tang and
                  Zhidian Huang and
                  Zhengxiao Du and
                  Xiao Liu and
                  Aohan Zeng and
                  Lei Hou and
                  Yuxiao Dong and
                  Jie Tang and
                  Juanzi Li},
  editor       = {Lun{-}Wei Ku and
                  Andre Martins and
                  Vivek Srikumar},
  title        = {LongBench: {A} Bilingual, Multitask Benchmark for Long Context Understanding},
  booktitle    = {Proceedings of the 62nd Annual Meeting of the Association for Computational
                  Linguistics (Volume 1: Long Papers), {ACL} 2024, Bangkok, Thailand,
                  August 11-16, 2024},
  pages        = {3119--3137},
  publisher    = {Association for Computational Linguistics},
  year         = {2024},
  url          = {https://doi.org/10.18653/v1/2024.acl-long.172},
  doi          = {10.18653/V1/2024.ACL-LONG.172},
  bibsource    = {dblp computer science bibliography, https://dblp.org}
}

@article{DBLP:journals/corr/abs-2506-01062,
  author       = {Thinh Pham and
                  Nguyen Nguyen and
                  Pratibha Zunjare and
                  Weiyuan Chen and
                  Yu{-}Min Tseng and
                  Tu Vu},
  title        = {SealQA: Raising the Bar for Reasoning in Search-Augmented Language
                  Models},
  journal      = {CoRR},
  volume       = {abs/2506.01062},
  year         = {2025},
  url          = {https://doi.org/10.48550/arXiv.2506.01062},
  doi          = {10.48550/ARXIV.2506.01062},
  eprinttype    = {arXiv},
  eprint       = {2506.01062},
  bibsource    = {dblp computer science bibliography, https://dblp.org}
}

@inproceedings{DBLP:conf/acl/JiangWL0L0Q24,
  author       = {Huiqiang Jiang and
                  Qianhui Wu and
                  Xufang Luo and
                  Dongsheng Li and
                  Chin{-}Yew Lin and
                  Yuqing Yang and
                  Lili Qiu},
  editor       = {Lun{-}Wei Ku and
                  Andre Martins and
                  Vivek Srikumar},
  title        = {LongLLMLingua: Accelerating and Enhancing LLMs in Long Context Scenarios
                  via Prompt Compression},
  booktitle    = {Proceedings of the 62nd Annual Meeting of the Association for Computational
                  Linguistics (Volume 1: Long Papers), {ACL} 2024, Bangkok, Thailand,
                  August 11-16, 2024},
  pages        = {1658--1677},
  publisher    = {Association for Computational Linguistics},
  year         = {2024},
  url          = {https://doi.org/10.18653/v1/2024.acl-long.91},
  doi          = {10.18653/V1/2024.ACL-LONG.91},
  bibsource    = {dblp computer science bibliography, https://dblp.org}
}

@inproceedings{gao2025how,
  title     = {How to Train Long-Context Language Models (Effectively)},
  author    = {Gao, Tianyu and Wettig, Alexander and Yen, Howard and Chen, Danqi},
  booktitle = {Proceedings of the 63rd Annual Meeting of the Association for Computational Linguistics (ACL)},
  year      = {2025},
}

@inproceedings{DBLP:conf/nips/AnML0LC24,
  author       = {Shengnan An and
                  Zexiong Ma and
                  Zeqi Lin and
                  Nanning Zheng and
                  Jian{-}Guang Lou and
                  Weizhu Chen},
  editor       = {Amir Globersons and
                  Lester Mackey and
                  Danielle Belgrave and
                  Angela Fan and
                  Ulrich Paquet and
                  Jakub M. Tomczak and
                  Cheng Zhang},
  title        = {Make Your {LLM} Fully Utilize the Context},
  booktitle    = {Advances in Neural Information Processing Systems 38: Annual Conference
                  on Neural Information Processing Systems 2024, NeurIPS 2024, Vancouver,
                  BC, Canada, December 10 - 15, 2024},
  year         = {2024},
  url          = {http://papers.nips.cc/paper\_files/paper/2024/hash/71c3451f6cd6a4f82bb822db25cea4fd-Abstract-Conference.html},
  bibsource    = {dblp computer science bibliography, https://dblp.org}
}

@article{jiang2023mistral,
  title   = {Mistral 7B},
  author  = {Jiang, Albert Q and Sablayrolles, Alexandre and Mensch, Arthur and Bamford, Chris and Chaplot, Devendra Singh and de las Casas, Diego and Bressand, Florian and Lengyel, Gianna and Lample, Guillaume and Saulnier, Lucile and others},
  journal = {arXiv preprint arXiv:2310.06825},
  year    = {2023},
  url     = {https://arxiv.org/abs/2310.06825}
}

@article{dubey2024llama,
  title   = {The Llama 3 Herd of Models},
  author  = {Dubey, Abhimanyu and Jauhri, Abhinav and Pandey, Abhinav and Kadian, Abhishek and Al-Dahle, Ahmad and Letman, Aiesha and Mathur, Akhil and Schelten, Alan and Yang, Amy and Fan, Angela and others},
  journal = {arXiv preprint arXiv:2407.21783},
  year    = {2024},
  url     = {https://arxiv.org/abs/2407.21783}
}

@inproceedings{DBLP:conf/iclr/HuSWALWWC22,
  author       = {Edward J. Hu and
                  Yelong Shen and
                  Phillip Wallis and
                  Zeyuan Allen{-}Zhu and
                  Yuanzhi Li and
                  Shean Wang and
                  Lu Wang and
                  Weizhu Chen},
  title        = {LoRA: Low-Rank Adaptation of Large Language Models},
  booktitle    = {The Tenth International Conference on Learning Representations, {ICLR}
                  2022, Virtual Event, April 25-29, 2022},
  publisher    = {OpenReview.net},
  year         = {2022},
  url          = {https://openreview.net/forum?id=nZeVKeeFYf9},
  bibsource    = {dblp computer science bibliography, https://dblp.org}
}

@inproceedings{DBLP:conf/iclr/LoshchilovH19,
  author       = {Ilya Loshchilov and
                  Frank Hutter},
  title        = {Decoupled Weight Decay Regularization},
  booktitle    = {7th International Conference on Learning Representations, {ICLR} 2019,
                  New Orleans, LA, USA, May 6-9, 2019},
  publisher    = {OpenReview.net},
  year         = {2019},
  url          = {https://openreview.net/forum?id=Bkg6RiCqY7},
  bibsource    = {dblp computer science bibliography, https://dblp.org}
}

@inproceedings{DBLP:conf/nips/ShaoHASDMZK24,
  author       = {Rulin Shao and
                  Jacqueline He and
                  Akari Asai and
                  Weijia Shi and
                  Tim Dettmers and
                  Sewon Min and
                  Luke Zettlemoyer and
                  Pang Wei Koh},
  editor       = {Amir Globersons and
                  Lester Mackey and
                  Danielle Belgrave and
                  Angela Fan and
                  Ulrich Paquet and
                  Jakub M. Tomczak and
                  Cheng Zhang},
  title        = {Scaling Retrieval-Based Language Models with a Trillion-Token Datastore},
  booktitle    = {Advances in Neural Information Processing Systems 38: Annual Conference
                  on Neural Information Processing Systems 2024, NeurIPS 2024, Vancouver,
                  BC, Canada, December 10 - 15, 2024},
  year         = {2024},
  url          = {http://papers.nips.cc/paper\_files/paper/2024/hash/a5d8aba27dfef4e849e8cb03fb87a954-Abstract-Conference.html},
  bibsource    = {dblp computer science bibliography, https://dblp.org}
}

@misc{Axolotl,
  author       = {Eric Winglian and Axolotl Contributors},
  title        = {{Axolotl}: Open‐Source LLM Fine‐Tuning Library},
  year         = {2025},
  url          = {https://github.com/axolotl-ai-cloud/axolotl},
  note         = {Accessed: 2025-07-21},
}

@inproceedings{DBLP:conf/emnlp/BaiLZHQH0DL24,
  author       = {Yushi Bai and
                  Xin Lv and
                  Jiajie Zhang and
                  Yuze He and
                  Ji Qi and
                  Lei Hou and
                  Jie Tang and
                  Yuxiao Dong and
                  Juanzi Li},
  editor       = {Yaser Al{-}Onaizan and
                  Mohit Bansal and
                  Yun{-}Nung Chen},
  title        = {LongAlign: {A} Recipe for Long Context Alignment of Large Language
                  Models},
  booktitle    = {Findings of the Association for Computational Linguistics: {EMNLP}
                  2024, Miami, Florida, USA, November 12-16, 2024},
  pages        = {1376--1395},
  publisher    = {Association for Computational Linguistics},
  year         = {2024},
  url          = {https://doi.org/10.18653/v1/2024.findings-emnlp.74},
  doi          = {10.18653/V1/2024.FINDINGS-EMNLP.74},
  bibsource    = {dblp computer science bibliography, https://dblp.org}
}

@inproceedings{DBLP:conf/acl/HwangCJSHP25,
  author       = {Taeho Hwang and
                  Sukmin Cho and
                  Soyeong Jeong and
                  Hoyun Song and
                  SeungYoon Han and
                  Jong C. Park},
  editor       = {Wanxiang Che and
                  Joyce Nabende and
                  Ekaterina Shutova and
                  Mohammad Taher Pilehvar},
  title        = {{EXIT:} Context-Aware Extractive Compression for Enhancing Retrieval-Augmented
                  Generation},
  booktitle    = {Findings of the Association for Computational Linguistics, {ACL} 2025,
                  Vienna, Austria, July 27 - August 1, 2025},
  pages        = {4895--4924},
  publisher    = {Association for Computational Linguistics},
  year         = {2025},
  url          = {https://aclanthology.org/2025.findings-acl.253/},
  bibsource    = {dblp computer science bibliography, https://dblp.org}
}

@inproceedings{CoLoR,
    title = "Efficient Long Context Language Model Retrieval with Compression",
    author = "Seo, Minju  and
      Baek, Jinheon  and
      Lee, Seongyun  and
      Hwang, Sung Ju",
    editor = "Che, Wanxiang  and
      Nabende, Joyce  and
      Shutova, Ekaterina  and
      Pilehvar, Mohammad Taher",
    booktitle = "Proceedings of the 63rd Annual Meeting of the Association for Computational Linguistics (Volume 1: Long Papers)",
    month = jul,
    year = "2025",
    address = "Vienna, Austria",
    publisher = "Association for Computational Linguistics",
    url = "https://aclanthology.org/2025.acl-long.740/",
    doi = "10.18653/v1/2025.acl-long.740",
    pages = "15251--15268",
    ISBN = "979-8-89176-251-0"
}

@article{narrativeqa,
    title = "The {N}arrative{QA} Reading Comprehension Challenge",
    author = "Ko{\v{c}}isk{\'y}, Tom{\'a}{\v{s}}  and
      Schwarz, Jonathan  and
      Blunsom, Phil  and
      Dyer, Chris  and
      Hermann, Karl Moritz  and
      Melis, G{\'a}bor  and
      Grefenstette, Edward",
    editor = "Lee, Lillian  and
      Johnson, Mark  and
      Toutanova, Kristina  and
      Roark, Brian",
    journal = "Transactions of the Association for Computational Linguistics",
    volume = "6",
    year = "2018",
    address = "Cambridge, MA",
    publisher = "MIT Press",
    url = "https://aclanthology.org/Q18-1023/",
    doi = "10.1162/tacl_a_00023",
    pages = "317--328"
}

@inproceedings{qasper,
    title = "A Dataset of Information-Seeking Questions and Answers Anchored in Research Papers",
    author = "Dasigi, Pradeep  and
      Lo, Kyle  and
      Beltagy, Iz  and
      Cohan, Arman  and
      Smith, Noah A.  and
      Gardner, Matt",
    editor = "Toutanova, Kristina  and
      Rumshisky, Anna  and
      Zettlemoyer, Luke  and
      Hakkani-Tur, Dilek  and
      Beltagy, Iz  and
      Bethard, Steven  and
      Cotterell, Ryan  and
      Chakraborty, Tanmoy  and
      Zhou, Yichao",
    booktitle = "Proceedings of the 2021 Conference of the North American Chapter of the Association for Computational Linguistics: Human Language Technologies",
    month = jun,
    year = "2021",
    address = "Online",
    publisher = "Association for Computational Linguistics",
    url = "https://aclanthology.org/2021.naacl-main.365/",
    doi = "10.18653/v1/2021.naacl-main.365",
    pages = "4599--4610"
}
